\documentclass[12pt]{amsart} 

\usepackage{amsmath,amssymb,latexsym,amsthm,graphicx,geometry}

\numberwithin{equation}{section}
\newcommand{\dd}{\mathrm{d}}
\newcommand{\Pm}{\mathbb{P}}
\newcommand{\R}{\mathbb{R}}

\newtheorem{lem}{Lemma}
\newtheorem{thm}{Theorem}
\newtheorem{prop}{Proposition}
\theoremstyle{definition}
\newtheorem{remark}{Remark}

\title{Posterior Bayesian Neural Networks with Dependent Weights} 

\author{Nicola Apollonio} 
\author{Giovanni Franzina}
\author{Giovanni Luca Torrisi}
\begin{document}
\maketitle



\begin{abstract}
We consider fully connected and feedforward deep neural networks with dependent and possibly heavy-tailed weights, as introduced in \cite{LAJLYC},  to address limitations of the standard Gaussian prior.  
It has been proved in \cite{LAJLYC} that, as the number of nodes in the hidden layers grows large,  according to a sequential and ordered limit, the law of the output converges weakly to a Gaussian mixture.  In this paper,  we study the neural network through the lens of the posterior distribution with a Gaussian likelihood.  If the random covariance matrix of the infinite-width limit is positive definite under the prior, we identify the posterior distribution of the output in the wide-width limit according to a sequential regime.  Remarkably,  we provide mild sufficient conditions to ensure the aforementioned invertibility of the random covariance matrix under the prior, thereby extending the results in \cite{CCMO}.  Among our results, we present sufficient conditions on some model parameters (the activation function and the associated Lévy measures) which ensure that the sequential limits are independent of the order.  We illustrate our findings with examples and numerical simulations.
\end{abstract}

{\scriptsize
\noindent{\em Keywords }{ Bayesian Analysis; Deep Neural Networks; Gaussian Mixture; Infinitely Divisible Distributions; Random Matrices} 

\noindent{\em 2020 Mathematics Subject Classification }{ 60F05; 62F15,68T07} 
}

\section{Introduction} 

A fully connected and feedforward neural network (which we simply refer to as a \lq\lq neural network\rq\rq\, throughout the paper) is the simplest neural network architecture and consists of a sequence of hidden layers stacked between an input layer and an output layer. Each node (or neuron) in a layer is connected to all the nodes in the subsequent layer. These networks are used to estimate unknown functions relating observed inputs to outputs.  Once the parameters of the network,  i.e.,  biases and weights,  have been estimated on the basis of a training dataset,  the network itself is a good approximation of the unknown target function.  
We refer the reader to \cite{GBC} for an introduction to this topic.
\paragraph{Overview.}\!\!\!The Bayesian approach to the analysis of neural networks allows to include in the model both a prior knowledge
on the parameters and the observed data through a prior distribution on neural network's parameters and a likelihood function,  respectively. 
Neal \cite{Neal1995,Neal1996} initiated the theoretical study of Bayesian neural networks by proving that, if a Bayesian shallow neural network is initialized with independent Gaussian parameters (i.e.,  the prior is Gaussian), then the output of the neural network converges in distribution to a Gaussian process as the number of neurons in the hidden layer increases,  namely,  in the infinite-width limit. This result was extended to deep neural networks two decades later~\cite{H,LBN18,MHR18} and has only recently been made quantitative using optimal transport theory~\cite{BT,T},  the Stein method for Gaussian approximation~\cite{ADFST,BGS,FHMNP},  and alternative techniques~\cite{BFF23,EMS21}.  Another promising approach to analyze Bayesian neural networks is through the lens of large deviations.  First results in this direction are given in \cite{HW,MPT}.  A different perspective is provided by the so-called mean field analysis of neural networks, see~\cite{HCG,MMN}.

The emergence of Gaussian processes improved our understanding of how large neural networks work and how to make them more efficient. It also motivated the use of Bayesian regression inference methods, see~\cite{LBN18}.  However, as noticed in \cite{Neal1995} and \cite{LAJLYC},  the connection with Gaussian processes  also highlighted the limitations of Bayesian neural networks with a Gaussian prior. Indeed,  there are at least three drawbacks with the choice of independent Gaussian weights: $(i)$ Hidden layers do not represent hidden features that capture important aspects of the data; $(ii)$ In the infinite-width limit the coordinates of the output become independent and identically distributed Gaussian processes,  which is usually undesirable; $(iii)$ The assumption of independent Gaussian weights is often unrealistic,  as estimated weights of deep neural networks show dependencies and heavy tail, see~\cite{Fortuin,MM19, WR}.  

To circumvent these limitations,  some authors~\cite{BR,FFP,JLLY} proposed priors which account for independent and identically distributed (i.i.d. for short) non-Gaussian weights.  However,  due to the independence assumption,  in the infinite-width limit,  the output of the neural network still converges to a stochastic process with independent coordinates.
A more structured prior on the weights has been proposed by \cite{APLL,LAJLYC}.  Letting \mbox{\scriptsize $W_{hj}^{(\ell+1)}$} denote the random weight between the node $h$ at the layer $\ell+1$ and the node $j$ at the layer $\ell$,  in \cite{LAJLYC} it is assumed that
\mbox{\scriptsize $W_{hj}^{(\ell+1)}:=
\sqrt{V_{n_{\ell},j}^{(\ell)}}N_{hj}^{(\ell)}$},
where \mbox{\scriptsize $V_{n_{\ell},j}^{(\ell)}$} are independent random variances,  which are identically distributed over $j$,  and \mbox{\scriptsize $N_{hj}^{(\ell)}$} are independent and Gaussian distributed random variables (r.v.'s for short),  with mean $0$ and variance $C_W>0$,  independent of the random variances. We refer the reader to Section~\ref{sec:model} for a rigorous description of the model. 
The prior proposed in \cite{LAJLYC} is more general than the Gaussian one,  which is retrieved setting \mbox{\scriptsize $V_{n_{\ell},j}^{(\ell)}$} $:=1/n_\ell$,  and it accounts for dependent and heavy-tailed weights. Indeed,  for fixed $\ell$ and $j$,  the weights \mbox{\scriptsize $\{W_{hj}^{(\ell+1)}\}_h$} are stochastically dependent and,  if the random variances are distributed according to the square of a heavy-tailed distribution with support on $(0,\infty)$,  a simple computation shows that the corresponding weights are also heavy-tailed.  

It is proved in \cite{LAJLYC} that if the aggregate random variances at the level of the layer $\ell+1$ converge in distribution 
(necessarily to an infinitely divisible law),  as the number of nodes in the layer $\ell$ grows large,  then the output of the neural network converges in law to a mixture of Gaussian processes whose coordinates are dependent but still identically distributed. Such a convergence takes place as the number of nodes in the hidden layers grows large,  according to a sequential limit with a prescribed order (see Theorem 16 in \cite{LAJLYC} or Theorem \ref{thm:starttraining} for the precise statement).  
In \cite{LAJLYC} it is left as challenging open problem to find more natural \lq\lq limiting schemes\rq\rq\, for the validity of such a convergence to a Gaussian mixture.  

To the best of our knowledge,  progress in the study of posterior Bayesian neural networks refers to models with a fixed variance for the Gaussian prior, see \cite{CP,  HBNPS, IV, PFP,T}. An exception is the recent work \cite{CFT},  where it is proved that,  if the parameters of the Bayesian neural network follow a Gaussian prior and the variance of both the last hidden layer and the Gaussian likelihood function is distributed according to an Inverse-Gamma prior law,  then the posterior Bayesian neural network converges to a Student-$t$ process in the infinite-width limit. 
\paragraph{Our contribution.}\!\!\!In this paper we are concerned with Bayesian neural networks with dependent and possibly heavy-tailed weights $\{W_{hj}^{(\ell+1)}\}$ as described above.  

In Theorem \ref{thm:04122025uno}, 
we establish sufficient conditions for the sequential limit of \cite[Theorem~16]{LAJLYC} to be independent of the order. 
As detailed in Appendix~\ref{AppA}, this generalization can be achieved via an alternate representation of the neural network, 
 which has been used in \cite{MPT} to study the large deviations of the output in the case of a Gaussian prior. 

As our main result, in Theorem~\ref{thm:bayesianlimit}, 
for Gaussian likelihoods we identify the infinite-width posterior distribution of the output and we show that, under a suitable probability measure,  it is a Gaussian mixture whose coordinates are dependent and not identically distributed.
This result entirely circumvents one of the limitations arising from the choice of a Gaussian prior. Such a result holds, however, under the assumption that the random covariance matrix of the infinite-width limit of the output is invertible under the prior distribution. 

Remarkably,  we provide sufficient conditions that guarantee such an invertibility assumption, by  extending the results in \cite{CCMO} to a more general framework,  see Theorems \ref{thm:09062025unobissone} and \ref{thm:09062025uno}.  These results are achieved by combining geometric measure-theoretic tools and an elementary,  but tricky,  linear algebraic construction.

The posterior law of the output of the Bayesian neural network at a finite width is very complicated,  especially for large values of the depth. One must resort to simulations to gain insight into this distribution. In the final section of the paper,  we provide some numerical illustrations of our main result.

\paragraph{Plan of the paper}
The paper is organized as follows.  Section~\ref{sec:1} presents the Bayesian neural network model which is investigated in the article. Section~\ref{sec:preliminaries} is devoted to preliminaries on matrices,  infinitely divisible distributions and mixtures of Gaussian distributions.  
Section~\ref{sec:widewidthprior} first presents the main result in \cite{LAJLYC}, i.e.,  the Gaussian mixture approximation of the output of the prior Bayesian neural network in the infinite-width limit,  according to a sequential limit with a prescribed order.  It then presents a generalization of this result, providing sufficient conditions to guarantee that the order does not matter in the sequential limit. As the proof of this result is quite technical, we have deferred it to the Appendix. Section~\ref{sec:widewidthposterior} studies the infinite-width limit of the posterior distribution of the output of the Bayesian neural network, providing identification of the limit in the case of a Gaussian likelihood. Section~\ref{sec:invertibility} addresses the challenging problem of invertibility of the prior random covariance matrix mentioned previously.  A couple of numerical illustrations of our main result are provided in Section~\ref{sec:8}. 

\section{Neural networks}\label{sec:1}

In this paper, for positive integers $p$ and $q$, $\mathbb R^p$ is the real numerical vector space whose elements are the column vectors with $p$ entries, while $\mathbb{R}^{p\times q}$ is the real vector space of real $p\times q$ matrices. The transposition operator $\top$ acts on $\mathbb R^p$ and $\mathbb{R}^{p\times q}$.  
Hereafter,  we also use the notation $X\stackrel{\mathcal{L}}{=}Y$ to denote that two r.v.'s $X$ and $Y$ have the same law and the notation $\overset{\mathcal{L}}{\to}$ to denote convergence in distribution.  In the sequel,  for $n\in\mathbb N^*:=\{1,2,\ldots\}$
we set $[n]:=\{1,2,\ldots,n\}$. 

\subsection{Neural networks and statistical learning}\label{sec:model0}

From a mathematical point of view,  fully connected and feedforward neural networks are a parametrized family of functions,  recursively defined as follows.  Let $L$, $n_0$,\ldots, $n_{L+1}\in\mathbb N^*$ be integers.  For $\ell\in[L]$,  we set
\begin{align}
Z_h^{(\ell+1)}(x)&:=B_h^{(\ell+1)}+\sum_{j=1}^{n_\ell}W_{hj}^{(\ell+1)}\sigma(Z_j^{(\ell)}(x)),\quad\text{$h\in[n_{\ell+1}]$}\nonumber\\
Z_h^{(1)}(x)&:=B_h^{(1)}+\sum_{j=1}^{n_0}W_{hj}^{(1)}x_j,\quad\text{$h\in[n_{1}]$}\nonumber
\end{align}
where $x\in\mathbb{R}^{n_0}$
is the input,  $\sigma:\mathbb R\to\mathbb R$ (a measurable function) is the activation function, 
$\{B_h^{(\ell)}\}_{\ell\in[L+1];h\in[n_\ell]}$ and $\{W_{hj}^{(\ell)}\}_{\ell\in[L+1];h\in[n_\ell];j\in[n_{\ell-1}]}$ are real parameters called biases and weights,  respectively.  The network consists of $L$ hidden layers stacked between the input layer (the layer $0$) and the output layer
(the layer $L+1$).  On each hidden layer $\ell\in[L]$,  there are $n_\ell$ nodes.  The network is called deep if $L\geq 2$ and shallow if $L=1$.  Throughout the paper,  we often apply $\sigma$ to vectors,  meaning that $\sigma(x)=(\sigma(x_1),\ldots,\sigma(x_{p}))^\top$ for $x=(x_1,\ldots,x_{p})^\top$.

In statistical learning,  neural networks are used to estimate unknown target functions.  
More precisely,  for a fixed network architecture ($L$,  $n_0,\ldots,n_{L+1}$,  $\sigma$),  for a given unknown target function $f:\mathbb R^{n_0}\to\mathbb R^{n_{L+1}}$ and a training dataset 
$\mathcal{D}:=\{(x(i),y(i))\}_{i\in[d]}\subset\mathbb{R}^{n_0}\times\mathbb{R}^{n_{L+1}}$,  $d\in\mathbb N^*$,  (i.e.,  couples of data $x(i)$ and outcomes $y(i):=f(x(i))$) the objective is to produce an estimate of the parameter 
$\Theta=(\{B_h^{(\ell)}\},\{W_{hj}^{(\ell)}\})$,  say $\widehat\Theta$,  in such a way that the output of the neural network is a good estimate of $f$,  i.e.,  
\[
(Z_1^{(L+1)}(x),\ldots,Z_{n_{L+1}}^{(L+1)}(x))^\top\Big|_{\Theta=\widehat\Theta}\approx y,
\quad\text{$\forall$ $(x,y)\in\mathcal D\cup\mathcal T$}
\]
where
$
\text{$\mathcal T:=\{(x'(j),y'(j))\}_{j\in[t]}\subset\mathbb{R}^{n_0}\times\mathbb{R}^{n_{L+1}}$,  $t\in\mathbb N^*$}
$
is a test set.

Throughout the paper we encode the inputs (data) in the $n_0\times d$ matrix $\bold x:=(x(1) \ldots x(d))$ and the outcomes (responses) in the $n_{L+1}\times d$ matrix 
$\bold y:=(y(1)\ldots y(d))$.

\subsection{Neural networks with dependent and possibly heavy-tailed weights,  and Bayesian statistical learning}\label{sec:model}

From now on,  all the random quantities are defined on a measurable space $(\Omega,\mathcal F)$,  on which
different probability laws will be defined.  Hereon $\mathcal{N}_1(\mu,v)$ denotes the one-dimensional Gaussian law with mean $\mu$ and variance $v$.  

The prior knowledge on the parameter $\Theta$ is modeled via a prior probability measure $\mathbb P_{{\rm prior}}$ on $(\Omega,\mathcal F)$.  In particular,
throughout the paper we assume that,  under $\mathbb P_{{\rm prior}}$,
 \begin{itemize}
\item For $\ell\in[L+1]$ and $h\in[n_\ell]$,  $B_h^{(\ell)}$ are r.v.'s with  
$
B_h^{(\ell)}\sim\mathcal{N}_1(0,C_B),
$
for a constant $C_B\geq 0$;

\item For $\ell\in[L+1]$,  $h\in[n_\ell]$ and $j\in[n_{\ell-1}]$, 
$W_{hj}^{(\ell)}$ are r.v.'s defined by 
$
W_{hj}^{(\ell)}:=\sqrt{V_{n_{\ell-1},j}^{(\ell-1)}}N_{hj}^{(\ell)},
$
where: for $j\in[n_0]$,  $V_{n_{0},j}^{(0)}:=n_0^{-1}$,  for $\ell=2,\ldots,L+1$,  $\{V_{n_{\ell-1},j}^{(\ell-1)}\}_{j\in[n_{\ell-1}]}$ are non-negative and identically distributed (over $j$) r.v.'s,  and
for $\ell\in[L+1]$,  $h\in[n_\ell]$ and $j\in[n_{\ell-1}]$, 
$N_{hj}^{(\ell)}$ are r.v.'s with
$
N_{hj}^{(\ell)}\sim\mathcal{N}_1(0,C_W),
$
for a constant $C_W>0$;

\item All the r.v.'s $\{B_h^{(\ell)},V_{n_{\ell-1},j}^{(\ell-1)},N_{hj}^{(\ell)}\}$
are stochastically independent.

\end{itemize}

Under these distributional assumptions,  one speaks of neural network with dependent and possibly heavy-tailed weights,  see the seminal paper \cite{LAJLYC},  where this prior has been introduced.  Indeed,  note that,  for a fixed $\ell\in\{2,\ldots,L+1\}$ and $j\in[n_{\ell-1}]$,  the random weights $W_{1j}^{(\ell)},\ldots,W_{n_\ell j}^{(\ell)}$ are dependent,  and that,  if $V_{n_{\ell-1},j}^{(\ell-1)}$ is distributed as the square of a random variable with a heavy tail law with support on $(0,\infty)$,  then the distribution of $W_{hj}^{(\ell)}$ is heavy tail.  Note also that if $V_{n_{\ell-1},j}^{(\ell-1)}:=\frac{1}{n_{\ell-1}}$,  then we recover the usual Gaussian prior.
The Bayesian appraoch to the statistical learning problem allows to incorporate in the model the training dataset $\mathcal D$ through a likelihood function $\mathcal{L}(\mathcal D,\Theta)$.  Then the posterior knowledge on the parameter is summarized by the
posterior probability measure
\[
\mathrm{d}\mathbb{P}_{{\rm posterior}}:=\frac{\mathcal{L}(\mathcal D,\Theta)\mathrm{d}\mathbb P_{{\rm prior}}}{\mathbb E_{\rm prior}\mathcal{L}(\mathcal D,\Theta)},
\]
where $\mathbb E_{\rm prior}$ denotes the expectation under the prior probability measure and it is assumed
$
\mathbb E_{\rm prior}\mathcal{L}(\mathcal D,\Theta)>0.
$
If the law of $\Theta$ under the prior has density $\mathrm{p}_{{\rm prior}}(\cdot)$,  then the law of $\Theta$ under the posterior has
density
$
\mathrm{p}_{{\rm posterior}}(\theta)\propto\mathcal{L}(\mathcal D,\theta)\mathrm{p}_{{\rm prior}}(\theta),
$
and one estimates the parameter $\Theta$ by maximizing $\mathrm{p}_{{\rm posterior}}(\cdot)$ via an adequate variant of the Stochastic Gradient Descent (maximum a posteriori estimate),  see e.g.  \cite{Fortuin}.

\section{Preliminaries}\label{sec:preliminaries}

In this section we provide some preliminaries on matrices and infinitely divisible distributions,  and we give the formal definition of Gaussian mixture distribution.
As general references for the first two topics, we cite the monographs \cite{LT,MKB} and~\cite{SATO}, respectively.

\subsection{Matrices}
For positive integers $p$ and $q$, both $\mathbb{R}^{p\times q}$ and $\mathbb{R}^{pq}$ are Euclidean spaces when equipped with the Frobenius inner product $\langle\cdot,\cdot\rangle_F$ and the standard dot product $\langle\cdot,\cdot\rangle$ defined, respectively by $\langle \bold A,\bold B\rangle_F=\mathrm{Tr}(\bold A^\top \bold B)$, with $\mathrm{Tr}(\cdot)$ the trace  operator, and $\langle a,b\rangle=a^\top b$. We denote by $\|\cdot\|_F$ and $\|\cdot\|$ the Frobenius norm and the standard Euclidean norm, induced by $\langle\cdot,\cdot\rangle_F$ and $\langle\cdot,\cdot\rangle$, respectively. For $\bold A\in \mathbb{R}^{p\times q}$,  let ${\rm vec}(\bold A)$ be the column vector in $\mathbb{R}^{pq}$ obtained by stacking the columns of $\bold A$ on top of each other, starting from the leftmost column. Since for $\bold A\in\mathbb{R}^{p\times q}$ and $\bold B\in\mathbb R^{q\times p}$, it holds that
\begin{equation}\label{eq:27122024uno}
	\mathrm{Tr}(\bold A\bold B)=\mathrm{vec}(\bold A^\top)^\top\mathrm{vec}(\bold B),
\end{equation}
it follows that ${\rm vec}: \mathbb{R}^{p\times q}\rightarrow \mathbb{R}^{pq}$ is an isometry between the Euclidean spaces $(\mathbb{R}^{p\times q},\langle\cdot,\cdot\rangle_F)$ and $(\mathbb{R}^{pq},\langle\cdot,\cdot\rangle)$. In view of this bijection, we think of a random matrix $\bold U$ as a matrix-valued r.v.  whose law is induced by the r.v.  ${\rm vec}(\bold U)$ (via ${\rm vec}^{-1})$. Accordingly,  the characteristic function of the random matrix $\bold U\in \mathbb{R}^{p\times q}$ is given by
$
\phi(\boldsymbol{\theta};\bold U)=\mathbb E\left(e^{\bold i\langle \boldsymbol{\theta},\bold U\rangle_F}\right)=\mathbb E\left(e^{\bold i\mathrm{Tr}(\boldsymbol{\theta}^\top \bold U)}\right)
$
where $\bold i:=\sqrt{-1}$ and $\boldsymbol{\theta}\in\mathbb{R}^{p\times q}$. The Kronecker (also known as tensor) product of matrices $\bold A=(a_{ij})\in\mathbb{R}^{p\times q}$ and $\bold B\in\mathbb R^{r\times s}$,
is the $pr\times qs$ real matrix $\bold A\otimes\bold B=(a_{ij}\bold B)_{1\leq i\leq p,\,1\leq j\leq q}$. If $\bold A$ and $\bold B$ are invertible matrices, then $(\bold A\otimes\bold B)^{-1}=\bold A^{-1}\otimes\bold B^{-1}$. Moreover, if $\bold A,\,\bold B$ and $\bold C$ are matrices such that $\bold{ABC}$ exists, then $\mathrm{vec}(\bold{ABC})=(\bold C^\top\otimes\bold A)\mathrm{vec}(\bold B)$. Therefore, with $\bold {Id}$ the identity matrix of appropriate dimension, choosing $\bold C=\bold {Id}$ and $\bold A$ invertible, identity \eqref{eq:27122024uno} and the preceding properties yield the useful relations 
\begin{equation}\label{eq:new}
	\mathrm{vec}(\bold B)^\top(\bold{Id}\otimes \bold A)\mathrm{vec}(\bold B)=\mathrm{Tr}(\bold{B^\top A B})\,\,\text{and}\,\,\mathrm{vec}(\bold B)^\top(\bold{ Id}\otimes \bold A)^{-1}\mathrm{vec}(\bold B)=\mathrm{Tr}(\bold{B^\top A^{-1} B}).
\end{equation}
For later purposes,
we recall that,  for every symmetric positive semi-definite matrix $\bold{A}=(a_{rs})_{r,s\in[d]}$,  there exists a unique symmetric positive semi-definite matrix $\bold{A}^{\sharp}=(a_{rs}^{\sharp})_{r,s\in[d]}$ (the square-root of $\bold A$) such that $\bold A^{\sharp}\bold A^{\sharp}=\bold A$,  i.e., 
$a_{rs}=\sum_{j=1}^{d}a_{rj}^{\sharp}a_{js}^{\sharp}$,  $r,s\in[d]$.
For $\bold A\in\mathbb{R}^{p\times q}$,  we denote by $rk(\bold A)$ the rank of $\bold A$,  i.e.,  the number of linearly independent columns or rows within $\bold A$.
We denote by $0$ the null vector of $\mathbb R^p$,  by ${\rm diag}_r(a_1,\ldots,a_r)$, $r\in\mathbb N^*$,  a diagonal $r\times r$ real matrix with diagonal elements $a_1,\ldots,a_r$,  by $\bold{Id}_p$ the $p\times p$ identity matrix,   and we set $1_p:=(1,\ldots,1)^\top\in\mathbb R^p$, $p\in\mathbb N^*$. 
The following lemma can be easily proved.

\begin{lem}\label{le:rk1}
Let $a_1,\ldots,a_n\in\mathbb R^p$, with $n\geq p$,  be $p$-dimensional (column) vectors and let $c_1,\ldots,c_n$ be arbitrarily chosen positive numbers.  Then the $p\times p$ matrix
$\sum_{i=1}^{n} c_i a_i a_i^\top$ is positive definite if and only if the $p\times n$ matrix $(a_1,\ldots,a_n)$ has rank equal to $p$.
\end{lem}

\subsection{Infinitely divisible distributions}\label{sec:ID}

A real-valued r.v.  $X$
is said to have an infinitely divisible law if,  for each $n\in\mathbb{N}^*$,  there exist i.i.d. $\mathbb R$-valued r.v.'s  $X_{n1},\ldots,X_{nn}$ such that $X\stackrel{\mathcal L}{=}X_{n1}+\ldots+X_{nn}$.  
If $X$ is a non-negative r.v.  it turns out that $X$ has an infinitely divisible distribution if and only if 
there exists a couple $(a,\rho)$, 
where $a\geq 0$ is a non-negative constant and $\rho$ is a L\'evy measure on $(0,\infty)$,  i.e.,  a Borel measure on $(0,\infty)$ with
\[
\int_{(0,\infty)}\min\{1,x\}\rho(\mathrm{d}x)<\infty,
\]
such that $X$ has characteristic function of the form ${\rm e}^{\Psi(u)}$ with
$
\Psi(u)=\bold i ua+\int_{(0,\infty)}(\mathrm{e}^{\bold i ux}-1)\rho(\mathrm d x).
$
We write $X=\mathrm{ID}(a,\rho)$.

\subsection{Gaussian mixtures}

We denote by $\mathcal{N}_m(c,\bold C)$ the $m$-dimensional Gaussian distribution with mean $c\in\mathbb R^m$ and covariance matrix $\bold C$.  Let $\kappa$ be a r.v.  with values on $\mathbb R^m$ and
let $\bold K$ be a positive semi-definite and symmetric random matrix
with values on $\mathbb R^{m\times m}$.
We say that a r.v.  $X$,  defined on the probability space $(\Omega,\mathcal F,\mathbb P)$ and with values on $\mathbb R^m$,  has the Gaussian mixture distribution (under $\mathbb P$) with random mean $\kappa$ and random covariance matrix $\bold K$,  denoted by
$\mathcal{GM}_m^{\mathbb P}(\kappa,\bold{K})$,  if $X\,|\,(\kappa,\bold K)\sim\mathcal{N}_m(\kappa,\bold K)$.  Note that if $\mathbb{P}({\rm det}\bold K>0)=1$,  then $X$ has density (with respect to the Lebesgue measure) $\mathbb E[\varphi_{(\kappa,\bold K)}(\xi)]$,  $\xi\in\mathbb R^m$, 
being $\varphi_{(c,\bold C)}$ the density of $\mathcal{N}_m(c,\bold C)$.
In the definition of Gaussian mixture we explicited the dependence on the probability measure $\mathbb P$ (with an upper index) since in this paper we will work with Gaussian mixtures under different probability measures.

\section{The infinite-width limit under the prior}\label{sec:widewidthprior}

We start introducing some notation.  For $\ell\in[L+1]$ and $i\in[d]$,  we set
\[
Z^{(\ell)}(x(i)):=(Z_1^{(\ell)}(x(i)),\ldots,Z_{n_\ell}^{(\ell)}(x(i)))^\top,
\]
and we consider the $n_{\ell}\times d$ random matrix 
$
\bold{Z}^{(\ell)}(\bold x):=(Z^{(\ell)}(x(1))\ldots Z^{(\ell)}(x(d))).
$
For later purposes,  we vectorialize the $n_{\ell}\times d$ random matrix $\bold{Z}^{(\ell)}(\bold x)$ defining the $n_{\ell}d$-dimensional random vector 
$
Z^{(\ell)}(\bold x):={\mathrm vec}((\bold{Z}^{(\ell)}(\bold x))^\top),\quad\ell\in[L+1].
$
The following Theorem \ref{thm:starttraining} is
the main result in \cite{LAJLYC} (see Theorem 16 therein).  It extends to the case of dependent and possibly heavy-tailed weights the Gaussian behavior in the infinite-width limit of a neural network with a Gaussian prior, see the seminal paper \cite{Neal1996} and the more recent contributions \cite{H,LBN18,MHR18}.

\begin{thm}\label{thm:starttraining}
Assume that:\\
$(i)$ The activation function $\sigma$ is continuous and such that
$\forall$ $z\in\mathbb R$,  $|\sigma(z)|\leq a_1+a_2|z|^{a_3}$ for some positive constants $a_1,a_2,a_3>0$.\\
\noindent $(ii)$ $\forall$ $\ell\in[L]$,  $\sum_{j=1}^{n_\ell}V_{n_\ell,j}^{(\ell)}\to\mathrm{ID}(a^{(\ell)},\rho^{(\ell)})$ in distribution as $n_\ell\to\infty$,  for some non-negative constant $a^{(\ell)}$ and L\'evy measure $\rho^{(\ell)}(\cdot)$ on $(0,\infty)$.

Then,  under $\mathbb{P}_{\mathrm{prior}}$,  we have
\begin{equation}\label{eq:04122025uno}
\text{$\lim_{n_L\to\infty}\ldots\lim_{n_1\to\infty}Z^{(L+1)}(\bold x)=G^{(L+1)}(\bold x)\sim\mathcal{GM}_{n_{L+1}d}^{\mathbb{P}_{\rm prior}}(0,\bold{Id}_{n_{L+1}}\otimes\bold{K}^{(L+1)}(\bold x))$ in distribution.}
\end{equation}
Here, 
$\bold{K}^{(L+1)}(\bold x)$
is defined by the following stochastic recursion:
\begin{equation}\label{eq:15032025tre}
\bold{K}^{(1)}(\bold x):=(K^{(1)}(x(i),x(i')))_{1\leq i,i'\leq d},
\end{equation}
where
$
K^{(1)}(x(i),x(i')):=C_B+C_W\frac{x(i)^\top x(i')}{n_0},
$
and,  for $\ell=2,\ldots,L+1$,   
\begin{align}
&\bold{K}^{(\ell)}(\bold x)
:=C_B 1_d 1_d^\top+C_W\Biggl(a^{(\ell-1)}\mathbb{E}_{{\rm prior}}[\sigma(\zeta_1^{(\ell-1)}(\bold x))\sigma(\zeta_1^{(\ell-1)}(\bold x))^\top\,|\,\bold{K}^{(\ell-1)}(\bold x)]\nonumber\\
&\qquad\qquad\qquad\qquad\qquad\qquad\qquad\qquad
+\sum_{j=1}^{N_{\ell-1}((0,\infty))}T_j^{(\ell-1)}\sigma(\zeta_j^{(\ell-1)}(\bold x))\sigma(\zeta_j^{(\ell-1)}(\bold x))^\top\Biggr).\label{eq:14052025due}
\end{align}
Here,  $\{\zeta_j^{(1)}(\bold x)\}_{j\geq 1}$ is a sequence of i.i.d.  r.v.'s with
$\zeta_1^{(1)}(\bold x)\sim\mathcal{N}_d(0,\bold{K}^{(1)}(\bold x))$,  for $\ell=3,\ldots,L+1$,  given $\bold{K}^{(\ell-1)}(\bold x)$,
$\{\zeta_j^{(\ell-1)}(\bold x)\}_{j\geq 1}$ is a sequence of i.i.d.  r.v.'s with 
$\zeta_1^{(\ell-1)}(\bold x)\sim\mathcal{N}_{d}(0,\bold{K}^{(\ell-1)}(\bold x))$,  and,  for $\ell=2,\ldots,L+1$,
$\{T_j^{(\ell-1)}\}_{j=1,\ldots,N_{\ell-1}((0,\infty))}$ are the points of a Poisson process on $(0,\infty)$ with mean measure $\rho^{(\ell-1)}$,  independent of the sequence
$\{\zeta_j^{(\ell-1)}(\bold x)\}_{j\geq 1}$.
\end{thm}

Let us represent $G^{(L+1)}(\bold x)$ (the limiting random vector in \eqref{eq:04122025uno}) as 
\begin{equation}\label{eq;30122025uno}
G^{(L+1)}(\bold x):=\mathrm{vec}((\bold{G}^{(L+1)}(\bold x))^\top),  
\end{equation}
where $\bold G^{(L+1)}(\bold x)$ is the
$n_{L+1}\times d$ random matrix 
$
\bold G^{(L+1)}(\bold x):=(G^{(L+1)}(x(1))\ldots G^{(L+1)}(x(d))),
$
and
$
G^{(L+1)}(x(i)):=(G_1^{(L+1)}(x(i)),\ldots,G_{n_{L+1}}^{(L+1)}(x(i)))^\top,\quad i\in[d].
$
Note that,  under $\mathbb{P}_{{\rm prior}}$,  the rows of $\bold G^{(L+1)}(\bold x)$ are pairwise dependent and uncorrelated,  and identically distributed according to the Gaussian mixture law with mean $0$ and random covariance matrix $\bold{K}^{(L+1)}(\bold x)$.  In Section \ref{sec:widewidthposterior} we will see that if we consider the posterior distribution with a Gaussian likelihood,  then,  under a suitable probability measure,  the infinite-width limit of the posterior output has dependent and,  in general,  not identically distributed coordinates (see Theorem \ref{thm:bayesianlimit}).

The limit in \eqref{eq:04122025uno} not only is sequential,  but it has also a prescribed order. 
Theorem \ref{thm:04122025uno} provides sufficient conditions which guarantee that in the sequential limit the order doesn't matter.  Its proof is quite technical and postponed in Appendix.

\begin{thm}\label{thm:04122025uno}
Assume $L\geq 2$ and the condition $(ii)$ of Theorem \ref{thm:starttraining}. If either
\begin{equation}\label{eq:05122025uno}
\text{$\sigma$ is continuous and bounded}
\end{equation}
or
\begin{equation}\label{eq:05122025due}
\text{$\sigma$ satisfies the condition $(i)$ of Theorem \ref{thm:starttraining} and $\rho^{(\ell)}((0,\infty))<\infty$ $\forall$ $\ell=2,\ldots,L$, }
\end{equation}
then,  for any fixed permutation $\pi$ of $1,\ldots,L$,  under $\mathbb{P}_{\rm prior}$,  
\[
\text{$\pi\lim Z^{(L+1)}(\bold x)=G^{(L+1)}(\bold x)\sim\mathcal{GM}_{n_{L+1}d}^{\mathbb{P}_{\rm prior}}(0,\bold{Id}_{n_{L+1}}\otimes\bold{K}^{(L+1)}(\bold x))$ in distribution} 
\]
i.e.,  in \eqref{eq:04122025uno} one can replace $\lim_{n_L\to\infty}\lim_{n_{L-1}\to\infty}\ldots\lim_{n_{1}\to\infty}$
with 
\[
\pi\lim:=\lim_{n_{\pi(L)}\to\infty}\lim_{n_{\pi(L-1)}\to\infty}\ldots\lim_{n_{\pi(1)}\to\infty}.
\]
\end{thm}

\section{The infinite-width limit under the posterior}\label{sec:widewidthposterior}

Theorem \ref{thm:starttraining} provides the distribution of the output of the neural network in the infinite-width limit,  under the prior probability measure $\mathbb P_{{\rm prior}}$ defined in Section \ref{sec:model}.  Here we give the distribution of the output of the neural network in the infinite-width limit,  under the posterior probability measure $\mathbb P_{{\rm posterior}}$ defined by
\begin{equation}\label{eq:05122025mattinauno}
\mathrm d\mathbb{P}_{{\rm posterior}}\propto\exp\left(-\sum_{i=1}^{d}\|Z^{(L+1)}(x(i))-y(i)\|^2\right)\mathrm d\mathbb P_{{\rm prior}}.
\end{equation}
Hereon,  (when the inverse matrices exist) we will consider the $n_{L+1}d\times n_{L+1}d$ 
random matrix
$
\boldsymbol{\Lambda}^{(L+1)}(\bold x):=\bold{Id}_{n_{L+1}}\otimes\bold{L}^{(L+1)}(\bold x)^{-1},
$
where $\bold L^{(L+1)}(\bold x)$ is the $d\times d$ random matrix
$
\bold L^{(L+1)}(\bold x):=2\bold{Id}_d+\bold{K}^{(L+1)}(\bold x)^{-1}.
$
We will also consider the $n_{L+1}d$-dimensional r.v.
$
\lambda^{(L+1)}(\bold x,\bold y):={\rm vec}(\boldsymbol\lambda^{(L+1)}(\bold x,\bold y)^\top),  
$
where $\boldsymbol\lambda^{(L+1)}(\bold x,\bold y)$ is the $n_{L+1}\times d$ random matrix 
$
\boldsymbol\lambda^{(L+1)}(\bold x,\bold y):=2\bold y\bold L^{(L+1)}(\bold x)^{-1}.
$
Hereafter,  for a $p$-dimensional r.v.  $X$ defined on a probability space $(\Omega,\mathcal F,\mathbb P)$ and with values on some metric space,  we denote by $\mathbb P_X$ (or sometimes by $(\mathbb P)_X$) the probability law induced by $X$ on the metric space.

The following theorem holds.

\begin{thm}\label{thm:bayesianlimit}
Assume conditions $(i)$ and $(ii)$ of Theorem \ref{thm:starttraining} and
\begin{equation}\label{eq:04032025tre}
\mathbb{P}_{{\rm prior}}(\mathrm{det}(\bold{K}^{(L+1)}(\bold x))>0)=1.
\end{equation}
Then
\begin{equation}\label{eq:05122025diecibissone}
\lim_{n_L\to\infty}\ldots\lim_{n_1\to\infty}(\mathbb{P}_{{\rm posterior}})_{Z^{(L+1)}(\bold x)}=\mathcal{GM}_{n_{L+1}d}^{\mathbb{S}}(\lambda^{(L+1)}(\bold x,\bold y),\boldsymbol\Lambda^{(L+1)}(\bold x))\quad\text{weakly.}
\end{equation}
Here $\mathbb{S}$ denotes the probability measure on $(\Omega,\mathcal F)$ defined by
\[
{\rm d}\mathbb{S}\propto\frac{
\exp\left(\mathrm{Tr}(\bold y(\bold{Id}_d+(2\bold{K}^{(L+1)}(\bold x))^{-1})^{-1})\bold y^\top\right)}{(\mathrm{det}(\bold{Id}_d+2\bold{K}^{(L+1)}(\bold x)))^{n_{L+1}/2}}\mathrm{d}\mathbb{P}_{{\rm prior}},
\]
and
$\mathcal{GM}_{n_{L+1}d}^{\mathbb{S}}(\lambda^{(L+1)}(\bold x,\bold y),\boldsymbol\Lambda^{(L+1)}(\bold x))$ is a Gaussian mixture under $\mathbb S$,  i.e.,  it has density 
\[
\mathbb{E}_{\mathbb S}[\varphi_{(\lambda^{(L+1)}(\bold x,\bold y),\boldsymbol\Lambda^{(L+1)}(\bold x))}(\xi)], \quad\text{$\xi\in\mathbb{R}^{n_{L+1}d}$}
\]
where $\mathbb{E}_{\mathbb S}$ denotes the mean with respect to $\mathbb S$.
\end{thm}

As clarified by Lemma \ref{thm:widewidthposteriorstar},  
the assumption \eqref{eq:04032025tre},  which
will be explored in the next section,  is crucial only to identify the limiting law.

\begin{remark}
Let $G^{(L+1)}(\bold x)$ be the r.v.  defined in 
\eqref{eq;30122025uno} and suppose that it is distributed according to the Gaussian mixture law in \eqref{eq:05122025diecibissone}.  Then,  under the probability measure $\mathbb{S}$, 
the rows of $\bold{G}^{(L+1)}(\bold x)$ are dependent and,  in general,  not identical distributed.  Indeed,  under $\mathbb S$,  the $i$-th row of $\bold{G}^{(L+1)}(\bold x)$ is distributed according to a Gaussian mixture with random covariance $\bold L^{(L+1)}(\bold x)^{-1}$ and random mean given by the $i$-th row of $2\bold y\bold L^{(L+1)}(\bold x)^{-1}$; note that the rows of this latter matrix are all equal if and only if the rows of $\bold y$ are such.
\end{remark}

\begin{remark}
Note that
\[
\mathbb{S}_{\bold{K}^{(L+1)}(\bold x)}({\rm d}\bold k)\propto\frac{
\exp\left(\mathrm{Tr}(\bold y(\bold{Id}_d+(2\bold{k})^{-1})^{-1})\bold y^\top\right)}{(\mathrm{det}(\bold{Id}_d+2\bold{k}))^{n_{L+1}/2}}(\mathbb{P}_{{\rm prior}})_{\bold{K}^{(L+1)}(\bold x)}({\rm d}\bold k),
\]
is a probability measure on the space of positive definite $d\times d$ symmetric matrices which can be interpreted as a posterior with likelihood 
\[
\frac{
\exp\left(\mathrm{Tr}(\bold y(\bold{Id}_d+(2\bold{k})^{-1})^{-1})\bold y^\top\right)}{(\mathrm{det}(\bold{Id}_d+2\bold{k}))^{n_{L+1}/2}}.
\]
Finally,  we remark that if $\bold{K}^{(L+1)}(\bold x)$ is deterministic (as it happens when the L\'evy measures are all equal to zero),  then $\mathbb S=\mathbb{P}_{\rm prior}$.
\end{remark}

\begin{remark}
A simple modification of the proof of Theorem \ref{thm:bayesianlimit} shows that if $L\geq 2$,  and we replace the condition $(i)$ of Theorem \ref{thm:starttraining} with either the assumption \eqref{eq:05122025uno} or the assumption \eqref{eq:05122025due},  then by Theorem \ref{thm:04122025uno} it follows that in \eqref{eq:05122025diecibissone} we can replace $\lim_{n_L\to\infty}\ldots\lim_{n_1\to\infty}$ with $\pi\lim$.
\end{remark}

The proof of Theorem \ref{thm:bayesianlimit} exploits the following Lemma \ref{thm:widewidthposteriorstar} which,  for the sake of completeness,  will be proved at the end of this section.  

Let $\{X_n\}_{n\geq 1}$ and $X$ be r.v.'s defined on the probability space $(\Omega,\mathcal F,\mathbb P)$ and with values on some metric space.  Consider on $(\Omega,\mathcal F)$ the tilted probability measures:
$
\mathrm d\mathbb{Q}_n\propto g(X_n)\mathrm d\mathbb P$ and
$\mathrm d\mathbb{Q}\propto g(X)\mathrm d\mathbb P
$
where $g$ is a bounded continuous function such that the normalizing constants are different from zero.

The following lemma holds.

\begin{lem}\label{thm:widewidthposteriorstar}
Suppose that $\mathbb{P}_{X_n}\to\mathbb{P}_X$ weakly,  as $n\to\infty$.  Then
$(\mathbb{Q}_{n})_{X_n}\to\mathbb{Q}_X$ weakly,  as $n\to\infty$. 
\end{lem}

\begin{proof}[Proof of Theorem \ref{thm:bayesianlimit}]
Thanks to Theorem \ref{thm:starttraining} and Lemma \ref{thm:widewidthposteriorstar}, we only need to compute the law $\mathbb{Q}_{G^{(L+1)}(\bold x)}$, where
\[
\mathrm d\mathbb{Q}\propto\exp\left(-\sum_{i=1}^{d}\|G^{(L+1)}(x(i))-y(i)\|^2\right)\mathrm d\mathbb{P}_{{\rm prior}}
\]
and
$
(\mathbb{P}_{{\rm prior}})_{G^{(L+1)}(\bold x)}=\mathcal{GM}_{n_{L+1}d}^{\mathbb{P}_{{\rm prior}}}(0,\bold{Id}_{n_{L+1}}\otimes\bold{K}^{(L+1)}(\bold x)).
$
Note that $\sum_{i=1}^{d}\|G^{(L+1)}(x(i))-y(i)\|^2=\|\bold{G}^{(L+1)}(\bold x)-\bold y\|^2_F$.  Let $\xi:=\mathrm{vec}(\boldsymbol\xi^\top)$,  where $\boldsymbol\xi:=[\xi(1)\ldots\xi(d)]$ and $\xi(i)\in\mathbb{R}^{n_{L+1}}$. By identity \eqref{eq:new} ad the definition of Frobenius norm, we have
\begin{align}
&\mathrm{d}\mathbb{Q}_{G^{(L+1)}(\bold x)}(\xi)=
\mathfrak c\exp\left(-\sum_{i=1}^{d}\|\xi(i)-y(i)\|^2\right)\nonumber\\
&\qquad
\times\mathbb E_{{\rm prior}}\Biggl[(\mathrm{det}(\bold{Id}_{n_{L+1}}\otimes\bold K^{(L+1)}(\bold x)))^{-1/2}
\exp\left(-\frac{1}{2}\xi^\top(\bold{Id}_{n_{L+1}}\otimes\bold{K}^{(L+1)}(\bold x))^{-1}\xi\right)\Biggr]\dd\xi\nonumber\\
&=
\mathfrak c\,\mathbb E_{{\rm prior}}\Biggl[\left(\frac{1}{\det \bold{K}^{(L+1)}(\bold x)}\right)^{\frac{n_{L+1}}{2}}\hspace{-1.7em}
\exp\left(-\mathrm{Tr}[(\boldsymbol\xi-\bold y)(\boldsymbol\xi-\bold y)^\top]-\frac{1}{2}\mathrm{Tr}[\boldsymbol\xi(\bold{K}^{(L+1)}(\bold x))^{-1}\boldsymbol\xi^\top]\right)\Biggr]\dd\xi\nonumber\\
&=\mathfrak c\,\mathbb E_{{\rm prior}}\Biggl[\left(\frac{1}{\det \bold{K}^{(L+1)}(\bold x)}\right)^{\frac{n_{L+1}}{2}}\hspace{-1.7em}
\exp\left(-\frac{1}{2}\mathrm{Tr}[2(\boldsymbol\xi-\bold y)(\boldsymbol\xi-\bold y)^\top-\boldsymbol\xi(\bold{K}^{(L+1)}(\bold x))^{-1}\boldsymbol\xi^\top]\right)\Biggr]\dd\xi
.\label{eq:04032025uno}
\end{align}
Next, for ease of notation set $\lambda=\lambda^{(L+1)}(\bold x,\bold y)$,  $\boldsymbol\lambda=\boldsymbol\lambda^{(L+1)}(\bold x,\bold y)$,  $\boldsymbol\Lambda=\boldsymbol\Lambda^{(L+1)}(\bold x)$,
$\bold{L}=\bold L^{(L+1)}(\bold x)$ and $\bold K=\bold{K}^{(L+1)}(\bold x)$ and observe that by identity \eqref{eq:new} we have
\begin{align}
(\xi-\lambda)^\top\boldsymbol\Lambda^{-1}(\xi-\lambda)
&=(\mathrm{vec}(\boldsymbol\xi^\top)-\mathrm{vec}(\boldsymbol\lambda^\top))^\top
(\bold{Id}_{n_{L+1}}\otimes\bold{L})
(\mathrm{vec}(\boldsymbol\xi^\top)-\mathrm{vec}(\boldsymbol\lambda^\top))
\nonumber\\
&=\mathrm{Tr}((\boldsymbol\xi-\boldsymbol\lambda)\bold{L}(\boldsymbol\xi-\boldsymbol\lambda)^\top).\label{eq:03032025dodici}
\end{align}
On the other hand
$
\mathrm{Tr}((\boldsymbol\xi-\boldsymbol\lambda)\bold{L}(\boldsymbol\xi-\boldsymbol\lambda)^\top)=\mathrm{Tr}(\boldsymbol\xi\bold{L}\boldsymbol\xi^\top)-2\mathrm{Tr}(\boldsymbol\lambda\bold{L}\boldsymbol\xi^\top)
+\mathrm{Tr}(\boldsymbol\lambda\bold{L}\boldsymbol\lambda^\top).
$
By definition $\boldsymbol\lambda^{(L+1)}=2\bold y\bold L^{-1}$. Hence $\boldsymbol\lambda\bold{L}\boldsymbol\xi^\top=
2\bold y\boldsymbol\xi^\top$. Plugging this expression in the previous identity and re-arranging in \eqref{eq:03032025dodici} yields
\begin{align}
{\rm Tr}[\boldsymbol\xi\bold{L}
\boldsymbol\xi^\top
-4\bold y\boldsymbol\xi^\top]
=(\xi-\lambda)^\top\boldsymbol\Lambda^{-1}(\xi-\lambda)-\mathrm{Tr}(\boldsymbol\lambda\bold{L}\boldsymbol\lambda^\top).
\label{eq:03032025trenta}
\end{align}
A straightforward computation shows
\begin{align}
{\rm Tr}[2(\boldsymbol\xi-\bold y)(\boldsymbol\xi-\bold y)^\top+\boldsymbol\xi\bold{K}^{-1}\boldsymbol\xi^\top]
={\rm Tr}[\boldsymbol\xi\bold{L}\boldsymbol\xi^\top
-4\boldsymbol\xi\bold y^\top]+2\mathrm{Tr}(\bold y\bold y^\top).\nonumber
\end{align}
On combining this latter relation with \eqref{eq:03032025trenta} we have
\begin{align}
{\rm Tr}[2(\boldsymbol\xi-\bold y)(\boldsymbol\xi-\bold y)^\top+\boldsymbol\xi\bold{K}^{-1}\boldsymbol\xi^\top]=
(\xi-\lambda)^\top\boldsymbol\Lambda^{-1}(\xi-\lambda)-\mathrm{Tr}(\boldsymbol\lambda\bold{L}\boldsymbol\lambda^\top)+2\mathrm{Tr}(\bold y\bold y^\top).\label{eq:20032025tre}
\end{align}
Plugging the right-hand side of \eqref{eq:20032025tre} in \eqref{eq:03032025dodici} yields
\begin{align}
\mathrm{d}\mathbb{Q}_{G^{(L+1)}(\bold x)}(\xi)
&=
\mathfrak c\mathbb E_{{\rm prior}}\Biggl[(\mathrm{det}\bold{K}^{(L+1)})^{-n_{L+1}/2}
\exp\left(\frac{1}{2}\mathrm{Tr}(\boldsymbol\lambda^{(L+1)}\bold{L}^{(L+1)}(\boldsymbol\lambda^{(L+1)})^\top)\right)
\nonumber\\
&\qquad\qquad\qquad
\times
\exp\left(-\frac{1}{2}(\xi-\lambda^{(L+1)})^\top(\boldsymbol\Lambda^{(L+1)})^{-1}(\xi-\lambda^{(L+1)})\right)
\Biggr]\dd\xi\nonumber\\
&=
\mathfrak c\mathbb E_{{\rm prior}}\Biggl[\frac{(\mathrm{det}\bold{K}^{(L+1)})^{-n_{L+1}/2}
\exp\left(\frac{1}{2}\mathrm{Tr}(\boldsymbol\lambda^{(L+1)}\bold{L}^{(L+1)}(\boldsymbol\lambda^{(L+1)})^\top)\right)}{(2\pi)^{-(n_{L+1}d)/2}\mathrm{det}(\boldsymbol\Lambda^{(L+1)})^{-1/2}}\nonumber\\
&\qquad\qquad
\times
\mathrm{det}(\boldsymbol\Lambda^{(L+1)})^{-1/2}\exp\left(-\frac{1}{2}(\xi-\lambda^{(L+1)})^\top(\boldsymbol\Lambda^{(L+1)})^{-1}(\xi-\lambda^{(L+1)})\right)
\Biggr]\dd\xi\nonumber\\
&=
\mathbb E_{\mathbb{S}}[\varphi_{(\lambda^{(L+1)},\boldsymbol\Lambda^{(L+1)})}(\xi)
]\dd\xi,\nonumber
\end{align}
and the proof is completed.
\end{proof}

\begin{proof}[Proof of Lemma \ref{thm:widewidthposteriorstar}] Set
$\mathfrak c_n:=(\mathbb E[g(X_n)])^{-1}$ and $\mathfrak c:=(\mathbb E[g(X)])^{-1}.$
Let $S$ be the metric space where the r.v.'s  $X_n$,  $n\geq 1$,  and $X$ take values and let $f:S\to\mathbb R$ be a bounded and continuous function.
We have
\begin{align}
&\int_{S}f(\xi)\mathrm{d}\mathbb{Q}_{n,X_n}(\xi)-\int_{S}f(\xi)\mathrm{d}\mathbb{Q}_{X}(\xi)
=\mathfrak c_n\int_{S}f(\xi)g(\xi)\mathrm{d}\mathbb{P}_{X_n}(\xi)-\mathfrak c\int_{S}f(\xi)g(\xi)\mathrm{d}\mathbb{P}_{X}(\xi)\nonumber\\
&=\mathfrak c_n\int_{S}f(\xi)g(\xi)(\mathrm{d}\mathbb{P}_{X_n}(\xi)-\mathrm{d}\mathbb{P}_{X}(\xi))
+(\mathfrak c_n-\mathfrak c)\int_{S}f(\xi)g(\xi)\mathrm{d}\mathbb{P}_{X}(\xi)
=:\mathfrak c_n\,I_n+(\mathfrak c_n-\mathfrak c)\,I.\nonumber
\end{align}
Therefore
\begin{align}
\Big|\int_{S}f(\xi)\mathrm{d}\mathbb{Q}_{n,X_n}(\xi)-\int_{S}f(\xi)\mathrm{d}& \mathbb{Q}_{X}(\xi)\Big|
\leq\mathfrak c_n\,|I_n|+|\mathfrak c_n-\mathfrak c|\,|I|
\leq\mathfrak c_n\,|I_n|+\|f\|_\infty\|g\|_\infty\,|\mathfrak c_n-\mathfrak c|.\nonumber
\end{align}
Since $g(\cdot)$ and $f(\cdot)g(\cdot)$ are bounded and continuous functions 
by the weak convergence of $\mathbb P_{X_n}$ to $\mathbb P_X$ it follows
$\lim_{n\to\infty}\mathfrak c_n=\mathfrak c$ and
$\lim_{n\to\infty}I_n=0$,  and the proof is completed.
\end{proof}

\section{Sufficient conditions for \eqref{eq:04032025tre}}\label{sec:invertibility}

Condition \eqref{eq:04032025tre} is crucial to identify the distribution of the infinite-width limit of the posterior output of the neural network (see Theorem \ref{thm:bayesianlimit}).  The following results provide sufficient conditions for \eqref{eq:04032025tre}.

\begin{thm}\label{thm:09062025unobissone}
Let the notation and assumption $(ii)$ of Theorem \ref{thm:starttraining} prevail.
Assume $\sigma$ Lipschitz continuous and non constant
and 
\begin{equation}\label{eq:05122025dieci}
\text{$a^{(\ell-1)}=0$ $\Rightarrow$ $\mathbb{P}_{{\rm prior}}(N_{\ell-1}((0,\infty))=\infty)=1$,  $\forall$ $\ell\in\{2,\ldots,L+1\}$.}
\end{equation}
If moreover
\begin{equation}\label{eq:datalinidp}
\text{The data $x(1),\ldots,x(d)$ are linearly independent vectors,  with $n_0\geq d$,}
\end{equation}
then \eqref{eq:04032025tre} holds.
\end{thm}

\begin{thm}\label{thm:09062025uno}
Let the notation and assumption $(ii)$ of Theorem \ref{thm:starttraining} prevail.
Assume $\sigma$ Lipschitz continuous and nonlinear
and
\begin{equation}\label{eq:09062025dieci}
\text{$a^{(1)}>0$ and $a^{(\ell-1)}=0$ $\Rightarrow$ $\mathbb{P}_{{\rm prior}}(N_{\ell-1}((0,\infty))=\infty)=1$,  $\forall$ $\ell\in\{3,\ldots,L+1\}$.}
\end{equation}
If moreover
\begin{equation}\label{eq:datadist}
\text{If $C_B>0$ then the data $x(1),\ldots,x(d)$ are all distinct}
\end{equation}
and
\begin{equation}\label{eq:datanotprop}
\text{If $C_B=0$ then the data $x(1),\ldots,x(d)$ are pairwise non-proportional,}
\end{equation}
then \eqref{eq:04032025tre} holds.
\end{thm}

The proof of these theorems relies on the following proposition which is proved later on in this section.

\begin{prop}\label{thm:detpos}
Let the notation and assumption $(ii)$ of Theorem \ref{thm:starttraining} prevail,
and assume  $\sigma$ Lipschitz continuous and non constant.
If for $\ell\in\{2,\ldots,L+1\}$ it holds
\begin{equation}\label{eq:09062025un}
\mathbb{P}_{{\rm prior}}(\mathrm{det}(\bold{K}^{(\ell-1)}(\bold x))>0)=1
\end{equation}
and
\begin{equation}\label{eq:06122025uno}
\text{$a^{(\ell-1)}=0$ $\Rightarrow$ $\mathbb{P}_{{\rm prior}}(N_{\ell-1}((0,\infty))=\infty)=1$, }
\end{equation}
then
\begin{equation}\label{eq:09062025uno}
\mathbb{P}_{{\rm prior}}(\mathrm{det}(\bold{K}^{(\ell)}(\bold x))>0)=1.
\end{equation}
\end{prop}

The proof of Proposition \ref{thm:detpos} exploits the following results,  which are proved at the end of the section.

\begin{prop}\label{prop:farea}
Let $\sigma$ be Lipschitz continuous and non constant,  and let $X$ be a $d$-dimensional r.v.  with a density with respect to the Lebesgue measure which is strictly positive on $\mathbb R^d$.  Then there exist $A=A_\sigma\in\mathcal{B}(\mathbb{R}^d)$ ($A$ does not depend on $X$) and $\varphi_{\sigma(X)}: A\to [0,\infty)$ measurable such that
\begin{equation}\label{eq:10092025uno}
\text{$\mathbb{P}(\sigma(X)\in C)=\int_{C}\varphi_{\sigma(X)}(y)\dd y$,  for any $C\in\mathcal{B}(A)$,  and $\mathbb{P}(\sigma(X)\in A)>0$.}
\end{equation}
\end{prop}

\begin{prop}\label{le:LebDec}
Let $X$ be a real-valued $m$-dimensional (column) r.v.  and suppose that its law and the Lebesgue measure on $\mathbb R^m$ are not singular and that $\mathbb E[XX^\top]<\infty$ (i.e.,  all the entries of the matrix $\mathbb E [XX^\top]$ are finite).  Then the matrix $\mathbb E[XX^\top]$ is positive definite.
\end{prop}

\begin{prop}\label{le:11052025uno}
Let $X_1,\ldots,X_r$ be $r$ (column) r.v.'s with values in $\mathbb R^q$,  $1\leq r\leq q$.  If the law of ${\rm vec}((X_1,\ldots,X_r))$ is absolutely continuous with respect to the Lebesgue measure on $\mathbb{R}^{qr}$,  then the random vectors $\{X_1,\ldots,X_r\}$ are linearly independent almost surely.
\end{prop}

\subsection{Proof of Theorems \ref{thm:09062025unobissone} and \ref{thm:09062025uno}}

\begin{proof}[Proof of Theorem \ref{thm:09062025unobissone}] The assumption \eqref{eq:datalinidp} implies $\mathrm{det}(\bold{K}^{(1)}(\bold x))>0$ (indeed,  it is easily verified that if the columns of a matrix $\bold A$ are linearly independent then the Gram matrix $\bold{A}^\top\bold A$ is positive definite).  The claim follows by Proposition
\ref{thm:detpos}.
\end{proof}

\begin{proof}[Proof of Theorem \ref{thm:09062025uno}] By Theorems 6 and 7 in \cite{CCMO} we have that the matrix 
\[
\mathbb{E}_{{\rm prior}}[\sigma(\zeta_1^{(1)}(\bold x))\sigma(\zeta_1^{(1)}(\bold x))^\top]
\] is positive definite.  Recall that
\begin{align*}
&\bold{K}^{(2)}(\bold x)
\nonumber\\
&
=C_B 1_d 1_d^\top+C_W\Biggl(a^{(1)}\mathbb{E}_{{\rm prior}}[\sigma(\zeta_1^{(1)}(\bold x))\sigma(\zeta_1^{(1)}(\bold x))^\top]
+\sum_{j=1}^{N_{1}((0,\infty))}T_j^{(1)}\sigma(\zeta_j^{(1)}(\bold x))\sigma(\zeta_j^{(1)}(\bold x))^\top\Biggr).
\end{align*}
Since $a^{(1)}>0$,  the matrix
$C_B 1_d 1_d^\top$ is positive semi-definite and the random matrix
\[
\text{$\sum_{j=1}^{\infty}T_j^{(1)}\sigma(\zeta_j^{(1)}(\bold x))\sigma(\zeta_j^{(1)}(\bold x))^\top$ is positive semi-definite, $\mathbb P_{{\rm prior}}$-a.s.}
\]
we have
$
\mathbb{P}_{{\rm prior}}(\mathrm{det}(\bold{K}^{(2)}(\bold x))>0)=1.
$
The claim follows by Proposition \ref{thm:detpos}.
\end{proof}

\subsection{Proof of Propositions \ref{thm:detpos},  \ref{prop:farea},   \ref{le:LebDec} and \ref{le:11052025uno}}

\begin{proof}[Proof of Proposition \ref{thm:detpos}] We assume $C_B>0$.  The proof in the case when $C_B=0$ is similar,  and therefore omitted.  Recall the definition of $\bold{K}^{(\ell)}(\bold x)$,  $\ell=2,\ldots,L+1$,  in the statement of Theorem \ref{thm:starttraining},  and note that it consists of the sum of three 
positive semi-definite matrices.  Since by assumption \eqref{eq:06122025uno} we have that $\mathbb{P}_{{\rm prior}}(N_{\ell-1}((0,\infty))=\infty)=1$ when $a^{(\ell-1)}=0$,  
the claim follows if we prove that
\begin{equation}\label{eq:15052025uno}
\text{$\mathbb{E}_{{\rm prior}}[\sigma(\zeta_1^{(\ell-1)}(\bold x))\sigma(\zeta_1^{(\ell-1)}(\bold x))^\top\,|\,\bold{K}^{(\ell-1)}(\bold x)]$ is positive definite,  $\mathbb{P}_{{\rm prior}}$-a.s.,}
\end{equation}
and
\begin{equation}\label{eq:10052025uno}
\text{$\mathrm{det}\left(\sum_{j=1}^{\infty}T_j^{(\ell-1)}\sigma(\zeta_j^{(\ell-1)}(\bold x))\sigma(\zeta_j^{(\ell-1)}(\bold x))^\top\right)>0$,\quad$\mathbb P_{{\rm prior}}$-a.s.}
\end{equation}
{\it Proof of \eqref{eq:15052025uno}.} By \eqref{eq:09062025un},  for $(\mathbb{P}_{{\rm prior}})_{\bold{K}^{(\ell-1)}(\bold x)}$-almost all $\bold k$,  under the prior,  $\zeta_1^{(\ell-1)}(\bold x)\,|\,\bold{K}^{(\ell-1)}(\bold x)=\bold k$
has a Gaussian density (which is obviously strictly positive on $\mathbb R^d$). So,  by Proposition \ref{prop:farea} we have that the law of
$\sigma(\zeta_1^{(\ell-1)}(\bold x))\,|\bold{K}^{(\ell-1)}(\bold x)=\bold k$ under $\mathbb{P}_{{\rm prior}}$ and the Lebesgue measure are not singular. 
By the Lipschitzianity of $\sigma$
and the Gaussianity of $\zeta_1^{(\ell-1)}(\bold x)\,|\,\bold{K}^{(\ell-1)}(\bold x)=\bold k$,
all the entries of 
\[
\mathbb{E}_{{\rm prior}}[\sigma(\zeta_1^{(\ell-1)}(\bold x))\sigma(\zeta_1^{(\ell-1)}(\bold x))^\top\,|\,\bold{K}^{(\ell-1)}(\bold x)=\bold k]
\]
are finite.  The claim then follows by Proposition \ref{le:LebDec}.\\
{\it Proof of \eqref{eq:10052025uno}. } Note that,  for $n\geq d$ arbitrarily fixed,  by Lemma \ref{le:rk1} we have
\begin{align}
&\Big\{rk((\sigma(\zeta_1^{(\ell-1)}(\bold x)),\ldots,\sigma(\zeta_n^{(\ell-1)}(\bold x))))=d\Big\}\nonumber\\
&\qquad\qquad
\equiv
\Big\{\sum_{j=1}^{n}T_j^{(\ell-1)}\sigma(\zeta_j^{(\ell-1)}(\bold x))\sigma(\zeta_j^{(\ell-1)}(\bold x))^\top\quad\text{is positive definite}\Big\}\nonumber\\
&\qquad\qquad
\subset
\Big\{\sum_{j\geq1}T_j^{(\ell-1)}\sigma(\zeta_j^{(\ell-1)}(\bold x))\sigma(\zeta_j^{(\ell-1)}(\bold x))^\top\quad\text{is positive definite}\Big\}.\nonumber
\end{align}
Therefore
\begin{align}
&\mathbb{P}_{{\rm prior}}\left(\mathrm{det}\left(\sum_{j\geq 1}T_j^{(\ell-1)}\sigma(\zeta_j^{(\ell-1)}(\bold x))\sigma(\zeta_j^{(\ell-1)}(\bold x))^\top\right)>0\right)\nonumber\\
&\quad\quad\qquad\geq\mathbb{P}_{{\rm prior}}\left(rk((\sigma(\zeta_1^{(\ell-1)}(\bold x)),\ldots,\sigma(\zeta_n^{(\ell-1)}(\bold x))))=d\right),\quad\text{for each $n\geq d$.}\label{eq:12052025uno}
\end{align}
By Proposition \ref{prop:farea},  for $(\mathbb{P}_{{\rm prior}})_{\bold{K}^{(\ell-1)}(\bold x)}$-almost all $\bold k$,
there exist $A=A_\sigma\in\mathcal{B}(\mathbb{R}^d)$ and\\ $\varphi_{\sigma(\zeta_1^{(\ell-1)}(\bold x))\,|\,\bold{K}^{(\ell-1)}(\bold x)=\bold k}: A\to [0,\infty)$ measurable such that
\begin{equation}\label{eq:06122025dieci}
\mathbb{P}_{{\rm prior}}(\sigma(\zeta_1^{(\ell-1)}(\bold x))\in C\,|\,\bold{K}^{(\ell-1)}(\bold x)=\bold k)=\int_C \varphi_{\sigma(\zeta_1^{(\ell-1)}(\bold x))\,|\,\bold{K}^{(\ell-1)}(\bold x)=\bold k}(y)\dd y,\,\,\text{$\forall$ $C\in\mathcal{B}(A)$} 
\end{equation}
(indeed,  since $\bold{K}^{(\ell-1)}(\bold x)$ is positive definite $\mathbb{P}_{{\rm prior}}$-a.s. ,  for $(\mathbb{P}_{{\rm prior}})_{\bold{K}^{(\ell-1)}(\bold x)}$-almost all $\bold k$,  the random vector
$\zeta_1^{(\ell-1)}(\bold x)\,|\,\bold{K}^{(\ell-1)}(\bold x)=\bold k$,  which has the Gaussian distribution $\mathcal{N}_d(0,\bold k)$,  has a strictly positive density).
Set
\begin{equation}\label{eq:11062025uno}
p_{\bold k}:=\mathbb{P}_{{\rm prior}}(\sigma(\zeta_1^{(\ell-1)}(\bold x))\in A\,|\,\bold{K}^{(\ell-1)}(\bold x)=\bold k)=\int_{A}\varphi_{\sigma(\zeta_1^{(\ell-1)}(\bold x))\,|\,\bold{K}^{(\ell-1)}(\bold x)=\bold k}(y)\dd y>0.
\end{equation}
For all $n\geq d$,  define the event
\[
D_{n,d}:=\{\text{the matrix $(\sigma(\zeta_1^{(\ell-1)}(\bold x)),\ldots,\sigma(\zeta_n^{(\ell-1)}(\bold x)))$ has at least $d$ columns in $A$}\}.
\]
We will prove later on that
\begin{equation}\label{06062025uno}
\mathbb{P}_{{\rm prior}}(D_{n,d}\cap\{rk((\sigma(\zeta_1^{(\ell-1)}(\bold x)),\ldots,\sigma(\zeta_n^{(\ell-1)}(\bold x))))\leq d-1\})=0.
\end{equation}
Thus,  by assuming \eqref{06062025uno}, 
\begin{align}
\mathbb{P}_{{\rm prior}}(D_{n,d})
&=\mathbb{P}_{{\rm prior}}(D_{n,d}\cap\{rk((\sigma(\zeta_1^{(\ell-1)}(\bold x)),\ldots,\sigma(\zeta_n^{(\ell-1)}(\bold x))))\geq d\})\nonumber\\
&\leq\mathbb{P}_{{\rm prior}}(rk((\sigma(\zeta_1^{(\ell-1)}(\bold x)),\ldots,\sigma(\zeta_n^{(\ell-1)}(\bold x)))))=d).\nonumber
\end{align}
So \eqref{eq:10052025uno} follows by \eqref{eq:12052025uno} and this latter relation if we prove that
$
\lim_{n\to\infty}\mathbb{P}_{{\rm prior}}(D_{n,d})=1.
$
Due to the dominated convergence theorem this latter limit holds if
$
\lim_{n\to\infty}\mathbb{P}_{{\rm prior}}(D_{n,d}\,|\,\bold{K}^{(\ell-1)}(\bold x)=\bold k)=1.
$ 
In order to prove this latter relation,  we note that since,  under $\mathbb{P}_{\rm prior}$, 
given $\{\bold{K}^{(\ell-1)}(\bold x)=\bold k\}$,
the r.v.'s $\sigma(\zeta_1^{(\ell-1)}(\bold x)),\ldots,\sigma(\zeta_n^{(\ell-1)}(\bold x))$ 
are i.i.d.,
recalling the definition of $p_{\bold k}$ in \eqref{eq:11062025uno},
we have that 
\begin{align}
\mathbb{P}_{{\rm prior}}(D_{n,d}\,|\,\bold{K}^{(\ell-1)}(\bold x)=\bold k)&=\mathbb{P}_{{\rm prior}}\Big(\big(\sigma(\zeta_1^{(\ell-1)}(\bold x),\ldots,\sigma(\zeta_n^{(\ell-1)}(\bold x))\big) \nonumber\\
& \qquad \qquad\qquad \qquad \text{has at least $d$ columns in $A$}\,|\,\bold{K}^{(\ell-1)}(\bold x)=\bold k\Big)\nonumber \\
&=\sum_{k=d}^{n}\binom{n}{k}p_{\bold k}a^k(1-p_{\bold k})^{n-k}.\nonumber
\end{align}
Since
$
\lim_{n\to\infty}\sum_{k=0}^{d}\binom{n}{k}p_{\bold k}^k(1-p_{\bold k})^{n-k}=0,
$
we have
\begin{align}
1=\lim_{n\to\infty}\sum_{k=0}^{n}\binom{n}{k}p_{\bold k}^k(1-p_{\bold k})^{n-k}& =\lim_{n\to\infty}\sum_{k=d}^{n}\binom{n}{k}p_{\bold k}^k(1-p_{\bold k})^{n-k}
=
\lim_{n\to\infty}\mathbb{P}_{{\rm prior}}(D_{n,d}\,|\,\bold{K}^{(\ell-1)}(\bold x)=\bold k),\nonumber
\end{align}
and the proof is completed.  It remains to prove \eqref{06062025uno}.  We start noticing that
\begin{align}
&D_{n,d}\cap\{rk((\sigma(\zeta_1^{(\ell-1)}(\bold x)),\ldots,\sigma(\zeta_n^{(\ell-1)}(\bold x))))\leq d-1\}\nonumber\\
&
\equiv\Biggl\{\text{$\exists$ $j\in\{d,\ldots,n\}$ and $i_1,\ldots,i_j\in[n]$ such that $\sigma(\zeta_{i_h}^{(\ell-1)}(\bold x))\in A$ $\forall$ $h\in[j]$}\nonumber\\
&\qquad\qquad\qquad\qquad\qquad
\text{and $rk((\sigma(\zeta_1^{(\ell-1)}(\bold x)),\ldots,\sigma(\zeta_n^{(\ell-1)}(\bold x))))\leq d-1$}\Biggr\}\nonumber\\
&
\subseteq\bigcup_{j=d}^{n}\bigcup_{\{i_1,\ldots,i_j\}\subseteq[n]}\Biggl\{
\text{
$\sigma(\zeta_{i_h}^{(\ell-1)}(\bold x))\in A$ $\forall$ $h\in[j]$}\nonumber\\
&\qquad\qquad\qquad\qquad\qquad
\text{and $rk((\sigma(\zeta_{i_1}^{(\ell-1)}(\bold x)),\ldots,\sigma(\zeta_{i_j}^{(\ell-1)}(\bold x))))\leq d-1$}\Biggr\}.\nonumber
\end{align}
Set
$
L_j:=\{(x_1,\ldots,x_j)\in\mathbb R^{d\times j}:\,\,rk((x_1,\ldots,x_j))\leq d-1\}.
$
Using the union bound,  it follows
\begin{align}
&\mathbb{P}_{{\rm prior}}(D_{n,d}\cap\{rk((\sigma(\zeta_1^{(\ell-1)}(\bold x)),\ldots,\sigma(\zeta_n^{(\ell-1)}(\bold x))))\leq d-1\})\nonumber\\
&\leq\sum_{j=d}^{n}\sum_{\{i_1,\ldots,i_j\}\subseteq[n]}\mathbb{P}_{{\rm prior}}\Biggl(\text{
$\sigma(\zeta_{i_h}^{(\ell-1)}(\bold x))\in A$ $\forall$ $h\in[j]$}\nonumber\\
&\qquad\qquad\qquad\qquad
\text{
and $rk(\sigma(\zeta_{i_1}^{(\ell-1)}(\bold x)),\ldots,\sigma(\zeta_{i_j}^{(\ell-1)}(\bold x)))\leq d-1$
}\Biggr)\nonumber\\
&=\sum_{j=d}^{n}\sum_{\{i_1,\ldots,i_j\}\subseteq[n]}\mathbb{P}_{{\rm prior}}((\sigma(\zeta_{i_1}^{(\ell-1)}(\bold x)),\ldots,\sigma(\zeta_{i_j}^{(\ell-1)}(\bold x)))\in A^j\cap L_j),\nonumber
\end{align}
and so it suffices to prove that
\begin{equation}\label{eq:04092025due}
\text{For each $j=d,\ldots,n$,  $\mathbb{P}_{{\rm prior}}((\sigma(\zeta_{i_1}^{(\ell-1)}(\bold x)),\ldots,\sigma(\zeta_{i_j}^{(\ell-1)}(\bold x)))\in A^j\cap L_j\,|\,\bold{K}^{(\ell-1)}(\bold x)=\bold k)=0.$}
\end{equation}
To this aim,  recall that,  under the prior,  given $\bold{K}^{(\ell-1)}(\bold x)$,  the r.v.'s $\zeta_{1}^{(\ell-1)}(\bold x),\ldots,\zeta_{n}^{(\ell-1)}(\bold x)$ are i.i.d and so,  using \eqref{eq:06122025dieci},  
we have that,  under the prior,  given $\bold{K}^{(\ell-1)}(\bold x)$,
the joint law of $\sigma(\zeta_{i_1}^{(\ell-1)}(\bold x)),\ldots,\sigma(\zeta_{i_j}^{(\ell-1)}(\bold x))$ is absolutely continuous with respect to the Lebesgue measure. Therefore, if $\bold{A}_j\in \R^{j\times d}$ is the matrix having these vectors as columns, then the rows of $\bold{A}_j$ satisfy the hypotheses of Proposition \ref{le:11052025uno} with $r=d$ and $q=j$. Hence $rk(\bold{A}_j)=d$, and we conclude that \eqref{eq:04092025due} holds.
\end{proof}

\begin{proof}[Proof of Proposition \ref{prop:farea}]
We will prove that \eqref{eq:10092025uno} holds with 
\[
A:=\mathbb{R}^d\setminus S\quad\text{and}\quad\varphi_{\sigma(X)}(y):=
\begin{cases}
\sum_{x\in\sigma^{-1}(y)}\frac{\varphi_X(x)}{J_\sigma(x)} & \text{if $\sigma^{-1}(y)$ is a countable set}\\
+\infty & \text{otherwise}
\end{cases}
\]
where $J_\sigma(x)$ is the absolute value of the determinant of the Jacobian matrix of $\sigma$,
\[
S:=\{y\in\mathbb R^d:\,\,\exists\,\, x\in\sigma^{-1}(y)\,\,\text{with}\,\, J_\sigma(x)=0\}
\] 
and $\varphi_X$ is the density of $X$.  Let $B\in\mathcal{B}(\R^d\setminus S)$ be arbitrarily fixed.  By the area formula applied to non-negative and measurable functions
$u$ (see Theorem 3.2.2 in \cite{fed}
and Theorem~2.71 in~\cite{AFP}, together with 
Eq.\ (2.47) thereafter) we have
\begin{equation}\label{eq:16092025quattro}
\int_{\R^d}u(x)J_\sigma(x)\dd x=\int_{\mathbb R^d}\sum_{x\in\sigma^{-1}(y)} u(x)\dd y.
\end{equation}
By taking
$
\text{$u(x):=\frac{\varphi_X(x)}{J_\sigma(x)}\bold{1}\{\sigma(x)\in B\}$,}
$
we get
\[
\mathbb{P}(\sigma(X)\in B)=\int_{\sigma^{-1}(B)}\varphi_X(x)\dd x=\int_B\varphi_{\sigma(X)}(y)\dd y.
\]
It remains to prove that $\mathbb{P}(\sigma(X)\in\mathbb{R}^d\setminus S)>0$.  This claim is equivalent to 
$\mathbb{P}(\sigma(X)\in S)<1$ i.e. $\mathbb{P}(X\in\mathbb R^d\setminus \sigma^{-1}(S))>0$.  Note that
\[
\text{$\sigma^{-1}(S)=T_0\cup T_1$,  where $T_0:=\{x\in\sigma^{-1}(S):\,\,J_\sigma(x)\neq 0\}$ and $T_1:=\{x\in\sigma^{-1}(S):\,\,J_\sigma(x)=0\}$.}
\]
We will show later on that
\begin{equation}\label{eq:16092025uno}
\mathrm{Leb}^{(d)}(T_0)=0,
\end{equation}
where $\mathrm{Leb}^{(d)}$ denotes the Lebesgue measure on $\mathbb R^d$.  Therefore, by the absolute continuity of $X$,  we have
$
\mathbb{P}(X\in\mathbb R^d\setminus \sigma^{-1}(S))=\mathbb{P}(X\in\mathbb R^d\setminus T_1),
$
and we reduce ourselves to prove that 
$\mathbb{P}(X\in\mathbb R^d\setminus T_1)>0$.
Since by assumption $X$ has density $\varphi_X$ such that $\varphi_X>0$ on $\mathbb R^d$ this latter inequality is equivalent to $\mathrm{Leb}^{(d)}(\mathbb{R}^d\setminus T_1)>0$.  In the final part of this proof,  we show that this latter claim is guaranteed by the fact that $\sigma$ is non constant and Lipschitz.  Reasoning by contradiction suppose that $\mathrm{Leb}^{(d)}(\mathbb{R}^d\setminus T_1)=0$,  
then for $\mathrm{Leb}^{(d)}$-almost all $x\in\mathbb R^d$ we have
$x\in T_1$ and so $J_\sigma(x)=0$.  
Since $\sigma$ acts componentwise on vectors,  the Jacobian matrix of $\sigma$ at $x=(x_1,\ldots,x_d)\in\mathbb{R}^d$ is diagonal with diagonal elements $\sigma'(x_i)$,  $i\in[d]$.  Therefore,  for $\mathrm{Leb}^{(1)}$-almost all $x\in\mathbb R$ it holds $\sigma'(x)=0$.  Since $\sigma$ is Lipschitz continuous this implies that $\sigma$ is constant,  which is a contradiction.  We finally prove \eqref{eq:16092025uno}.  By the area formula \eqref{eq:16092025quattro} with
$
u(x):=\frac{1}{J_\sigma(x)}\bold{1}\{x\in T_0\}
$
we have
\begin{align}
{\rm Leb}^{(d)}(T_0)&=\int_{\mathbb R^d}\sum_{x\in\sigma^{-1}(y)}\frac{1}{J_\sigma(x)}\bold{1}\{x\in T_0\}\dd y
\nonumber\\
&
=\int_{\sigma(T_0)}\sum_{x\in\sigma^{-1}(y)}\frac{1}{J_\sigma(x)}\dd y\le
\int_{S}\sum_{x\in\sigma^{-1}(y)}\frac{1}{J_\sigma(x)}\dd y=0,\nonumber
\end{align}
where the vanishing of the latter integral follows noticing that the Lipschitzianity of $\sigma$ implies ${\rm Leb}^{(d)}(S)=0$,  which,  in turn,
follows by the (co)area formula \eqref{eq:16092025quattro}
with the choice $u(x)=\bold{1}\{x\in T_1\}$ (this is the
weak Morse-Sard property, see Remark $(ii)$ after
Theorem 1 in~\cite[Section~3.4.2]{EG}).
\end{proof}

\begin{proof}[Proof of Proposition \ref{le:LebDec}]
We start noticing that the matrix $\mathbb E[XX^\top]$ is positive semi-definite.  
Reasoning by contradiction,  suppose that there exists $v\in\mathbb R^m\setminus\{0\}$ such that $\mathbb E|v^\top X|^2=0$.  Hence, $v^\top X=0$  $\mathbb P$-a.s.,  i.e. $\mathbb P_X(H_v)=1$,  where by $H_v$ we are  denoting the hyperplane
$\{y\in\mathbb R^m:\,\, v^\top y=0\}.$ Let $A^c$ denote the complement of a set $A$.  Since $\mathbb P_X$ and the Lebesgue measure are not singular,  letting $\mathbb{P}_{X,\ll}$ denote the continuous part in the Lebesgue decomposition of $\mathbb P_X$ with respect to the Lebesgue measure,  we have that there exists a Borel set $C\subseteq\mathbb R^m$ such that $\mathbb{P}_{X,\ll}(C)>0$.  Therefore,  since $\mathbb{P}_{X,\ll}(H_v)=0$,  we have
\begin{align*}
0=1-\mathbb{P}_X(H_v)=\mathbb{P}_X(H_v^c)\geq\mathbb{P}_{X,\ll}(H_v^c)&\geq\mathbb{P}_{X,\ll}(C\cap H_v^c)\nonumber\\
&=\mathbb{P}_{X,\ll}(C\cap H_v^c)+\mathbb{P}_{X,\ll}(B\cap H_v)=\mathbb{P}_{X,\ll}(C)>0.
\end{align*}
This is a contradiction,  and the proof is completed.
\end{proof}

\begin{proof}[Proof of Proposition \ref{le:11052025uno}]
Recall that,  for $r\in[q]$,  the vectors $a_1,\ldots,a_r$ of $\R^q$ are linearly dependent if and only if every $r\times r$ submatrix of $\bold{A}:=(a_1,\ldots,a_r)$ has 
determinant equal to zero.  Let $\mathcal{J}_r$ be the collection of subsets of $[q]$ with cardinality equal to $r$. 
For $J\in \mathcal{J}_r$, $J=\{j_1,\ldots j_r\}$, let $p_J: \R^q\rightarrow \R^r$ be the projection $(x_1,\ldots,x_q)^\top\mapsto (x_{j_1},\ldots,x_{j_r})^\top$. Clearly,  for every $J\in \mathcal{J}_r$,  $\widetilde{p}_J(\bold A):=(p_J(a_1),\ldots,p_J(a_r))$ 
is an $r\times r$ submatrix of $\bold A$,  and every $r\times r$ submatrix of $\bold A$ arises in this way for some $J\in \mathcal{J}_r$. 
Note that
$a_1,\ldots,a_r$ are linearly dependent if and only if $f(\bold A)=0$ where $f:\R^{q\times r}\rightarrow [0,\infty)$ is defined by $$f(\bold A):=\sum_{J\in\mathcal{J}_r}|{\rm det}(\widetilde{p}_J(\bold A))|.$$  
It is easily realized that $f$ is a continuous function (as composition of continuous functions),  and so $f$ is measurable.
Set $\bold{X}=(X_1,\ldots,X_r)$ and
$
E:=\{\text{the r.v.'s $X_1,\ldots,X_r$ are linearly independent}\}.
$
We have
$\Pm(E)=1-\mathbb{P}(f(\bold X)=0)\geq 1-\Pm({\rm det}(\widetilde{p}_{\{1,\ldots,r\}}(\bold{X}))=0).$
Since ${\rm det}(\widetilde{p}_{\{1,\ldots,r\}}(\cdot))$ is a polynomial function,
by the absolute continuity of the law of
${\rm vec}((X_1,\ldots,X_r))$ with respect to the Lebesgue measure we have
\[
\Pm({\rm det}(\widetilde{p}_{\{1,\ldots,r\}}(\bold{X}))=0)=0.
\]
Here, we also used
the classical fact that every non-identically zero polynomial function $p: \R^m\rightarrow \R$ is non-zero almost everywhere, 
which can be seen by induction on $m$ in view of Fubini's theorem (see, e.g.,  \cite[Section~2.6.5]{fed}). 
Therefore $\mathbb P(E)=1$,  and the proof is completed.
\end{proof}

\section{Numerical illustrations}\label{sec:8}

In this section we provide numerical illustrations of Theorem \ref{thm:bayesianlimit}.
More precisely,  we consider two models of neural networks with dependent
and heavy-tailed weights,  we sample from the posterior distribution of the output at a finite and at an infinite width and we verify the convergence,  as the number of nodes in the hidden layers grows large.  We start introducing the models.

\subsection{Model 1}\label{sec:22052025uno}

The first model that we analyze is a neural network with dependent weights,
as defined in Sections \ref{sec:model0} and \ref{sec:model},  with $C_B>0$ and
$V_{n_\ell,j}^{(\ell)}:= Y_j^{(\ell)}/n_\ell$,  $j\in[n_\ell]$,  $\ell\in[L]$,
where,  under $\mathbb P_{{\rm prior}}$,  $\{Y_j^{(\ell)}\}_{j\geq 1,\ell\in[L]}$ are independent r.v.'s and $\{Y_j^{(\ell)}\}_{j\geq 1}$ are identically distributed with $\mathbb P_{{\rm prior}}(Y_1^{(\ell)}>0)=1$ and $\mathbb E_{{\rm prior}} Y_1^{(\ell)}\in (0,\infty)$.  By the law of the large numbers,  for each $\ell\in[L]$,  we have
$
\sum_{j=1}^{n_\ell}V_{n_\ell,j}^{(\ell)}\to \mathbb E_{{\rm prior}} Y_1^{(\ell)},
$
$\mathbb P_{{\rm prior}}$-a.s.,  as $n_\ell\to\infty$.
Therefore,  the assumption $(ii)$ of Theorem \ref{thm:starttraining} holds with $a^{(\ell)}=\mathbb E_{{\rm prior}} Y_1^{(\ell)}$ and $\rho^{(\ell)}\equiv 0$.  
If the inputs are e.g.  distinct and the activation function $\sigma$ is e.g.  Lipschitz continuous and nonlinear,  then 
all the hypotheses of Theorem \ref{thm:09062025uno} are fullfilled and the corresponding assumption
\eqref{eq:04032025tre} holds.  To make the model more realistic,  under the prior,  the weights should be heavy-tailed (see the introduction and the references \cite{MM19,WR}).  For instance,  this happens if,  under $\mathbb{P}_{{\rm prior}}$,
$\{Y_j^{(\ell)}\}_{j\geq 1,\ell\in[L]}$ are i.i.d.  r.v.'s with $Y_1^{(1)}\stackrel{\mathcal L}{=}(WE)^2$,  where $WE$ has the Weibull distribution with parameters $(1,1/2)$,  i.e.,  it has density (with respect to the Lebesgue measure)
$
f_{WE}(x):=\frac{1}{2}x^{-1/2}\mathrm{e}^{-\sqrt{x}},\quad x>0
$
(we could work with other heavy-tailed distributions,  however,  since we are interested in simulating the model,  we prefer to make a specific choice of the parameters soon at this stage).
Indeed,  in such a case,  the Laplace transform of $W_{11}^{(2)}$ on $\mathbb{R}\setminus\{0\}$ is equal to infinity,  as a simple calculation shows.   
Note also that a standard computation gives
$\mathbb{E}_{{\rm prior}}(WE)^2=24$,  and so in this specific case $a^{(\ell)}=24$,  $\ell\in[L]$.

\subsection{Model 2}\label{sec:model2}

The second model that we analyze is a neural network with dependent weights,  as defined in Sections \ref{sec:model0} and \ref{sec:model},  with $C_B>0$ and $V_{n_\ell,j}^{(\ell)}:=\frac{\pi^2}{n_\ell^2}Y_j^{(\ell)}$,  $j\in[n_\ell]$,  $\ell\in[L]$, 
where,  under $\mathbb P_{{\rm prior}}$,  
$\{Y_j^{(\ell)}\}_{j\geq 1,\ell\in[L]}$ is a family of i.i.d.  r.v.'s with $Y_1^{(1)}\stackrel{d}{=}(HC)^2$,  where $HC$ denotes a r.v.  distributed according to the half-Cauchy law,  i.e.,  with probability
density (with respect to the Lebesgue measure)
$
f_{HC}(x):=\frac{2}{\pi(1+x^2)}\bold{1}\{x>0\}.
$
Note that $\mathbb P_{{\rm prior}}(Y_1^{(1)}>0)=1$,  $\ell\in[L]$.
Under $\mathbb{P}_{{\rm prior}}$,  it turns out (see Appendix E in \cite{LAJLYC}) 
that,  for $\ell\in[L]$,
$\sum_{j=1}^{n_\ell}V_{n_\ell,j}^{(\ell)}\to {\rm ID}(0,\rho),$
in distribution,  as $n_\ell\to\infty$,
where
\begin{equation}\label{eq:01072025uno}
\rho(\dd x):=x^{-3/2}\bold{1}\{x>0\}\dd x.
\end{equation}
Therefore,  the assumption $(ii)$ of Theorem \ref{thm:starttraining} holds with $a^{(\ell)}=0$ and $\rho^{(\ell)}=\rho$ for each $\ell$.
Note that,  for any $\varepsilon>0$,
$\rho((\varepsilon,\infty))=2\varepsilon^{-1/2}$ and $\rho((0,\varepsilon])=+\infty,$
and therefore a Poisson process with mean measure $\rho$ have infinitely many points on $(0,\infty)$.  Consequently,  the assumption \eqref{eq:05122025dieci} holds.  If the inputs are such that \eqref{eq:datalinidp} is satisfied and the
activation function $\sigma$ is Lipschitz and non constant,  then  all the hypotheses of Theorem \ref{thm:09062025unobissone} are fullfilled and the corresponding assumption
\eqref{eq:04032025tre} holds. 
Here again,  to make the model more realistic,  under the prior,  the weights should be heavy-tailed, see the introduction and \cite{MM19,WR}.  
This is the case for the Model 2.  Indeed,  the Laplace transform of $W_{11}^{(2)}$ on $\mathbb{R}\setminus\{0\}$ is equal to infinity,  as can be easily verified by a simple
computation.

\subsection{Simulation of Model 1}\label{ss:model1}

\begin{figure}
\centering
\includegraphics[width=12.5cm]{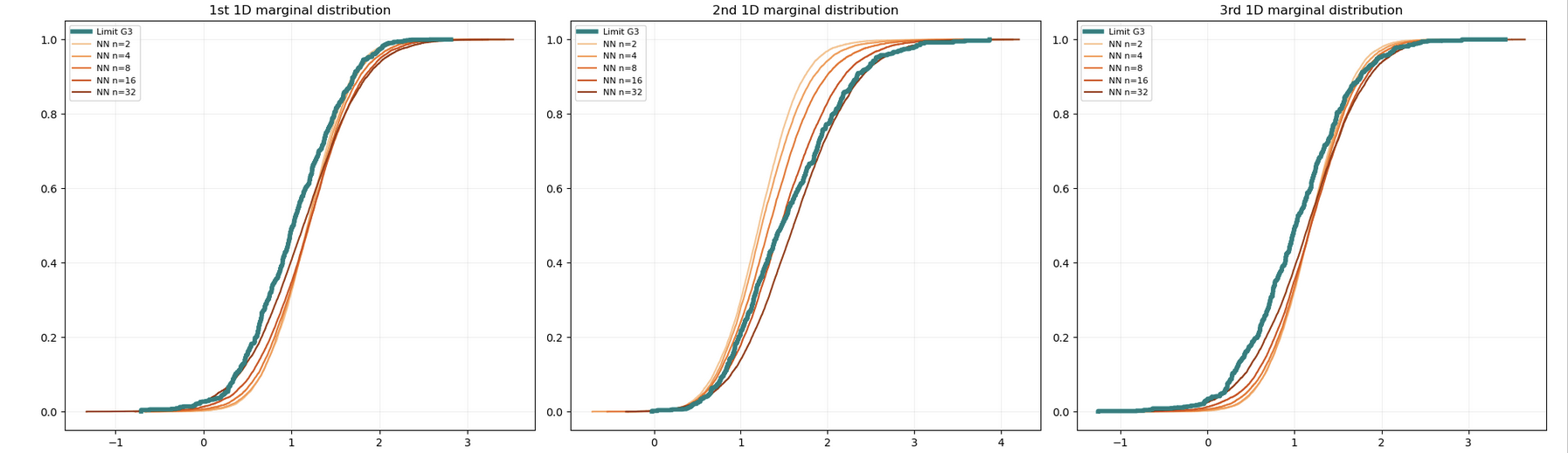}
\caption{
{\bf Simulation of Model 1}.  The non-green curves are estimates of the one dimensional marginal distribution functions (with $n_1=n$ nodes in the first hidden layer and $n_2=2n$
in the second hidden layer,  for different values 
of $n=4,8,16,32$) of the distribution function of the posterior law of the $3$-variate output $Z^{(3)}(\bold x)$.  The green curves are the one dimensional marginal distribution functions of the distribution function of the $3$-variate infinite-width limit $G^{(3)}(\bold x)$.  All the other parameters of the model are specified in Section~\ref{ss:model1}.}\label{fig1}
\end{figure}

We consider the Model 1 with $C_B:=1$,  $C_W:=1$,  $L:=2$,  $n_0:=4$,  $n_3:=1$, 
$n_1=n\in\{2,4,8,16,32\}$,  $n_2=2n$,  $d:=3$,
$x(1):=(1,0,0,0)^\top$,  $x(2):=(0,1,0,0)^\top$,  
$x(3):=(0,0,1,0)^\top$,
 $d:=3$,    $y(1):=f(x(1))$,  $y(2):=f(x(2))$,  $y(3):=f(x(3))$,  where 
 $f(v_1,v_2,v_3,v_4):= v_1^2+2v_1v_2+v_3^2$,  and 
activation function $\sigma(x):=\max\{0,x\}$.  Then
\begin{equation}\label{eq:09012026uno}
\bold{K}^{(1)}(\bold x):=(K^{(1)}(x(i),x(i')))_{1\leq i,i'\leq 3},\qquad
\text{where}\qquad
K^{(1)}(x(i),x(i')):=1+\frac{1}{4}x(i)^\top x(i'),
\end{equation}
and for $\ell=1,2$, 
given $\zeta_1^{(\ell)}(\bold x)\sim\mathcal{N}_3(0,\bold{K}^{(\ell)}(\bold x))$,
\begin{align*}
\bold{K}^{(\ell+1)}(\bold x)
=1_3 1_3^\top+24\mathbb{E}_{{\rm prior}}[\max\{\zeta_1^{(\ell )}(\bold x),0\}\max\{\zeta_1^{(\ell)}(\bold x),0\}^\top].
\end{align*} 
By Theorem \ref{thm:bayesianlimit} if we denote by $G^{(3)}(\bold x)$ the infinite-width limit of the posterior law of the output, then
$
G^{(3)}(\bold x)\sim\mathcal{N}_3(\lambda^{(3)}(\bold x,\bold y),\boldsymbol\Lambda^{(3)}(\bold x)).
$
Since,  for $A\in\mathcal{B}(\mathbb{R}^{3})$,
\[
\mathbb{P}_{{\rm posterior}}(Z^{(3)}(\bold x)\in A)\propto\mathbb{E}_{{\rm prior}}\left[\bold{1}_A(Z^{(3)}(\bold x))\exp\left(-\sum_{i=1}^{3}\|Z^{(3)}(x(i))-y(i)\|^2\right)\right],
\]
if the number of nodes is kept fixed,  we sample from the posterior law of 
$Z^{(3)}(\bold x)$
simulating this random variable under the prior,  and then using a Monte Carlo estimator.  See Figure~\ref{fig1} for a numerical validation of Theorem \ref{thm:bayesianlimit}. 

\subsection{Simulation of Model 2}\label{ss:model2}

We consider the Model 2 with $C_B:=1$,  $C_W:=1$,  $L=1$,  $n_0:=4$,  $n_1\in\{2,4,8,16,32\}$,  $n_2:=1$,  $d:=3$,  
$x(1):=(1,0,0,0)^\top$,  $x(2):=(0,1,0,0)^\top$,  $x(3):=(0,0,1,0)^\top$,  $y(1):=f(x(1))$,  $y(2):=f(x(2))$,  $y(3):=f(x(3))$,  where $f(v_1,v_2,v_3,v_4)=v_1^2+2v_1v_2+v_3^2$,  and
activation function $\sigma(x):=\max\{0,x\}$.
Then $\bold{K}^{(1)}(\bold x)$ is given by
\eqref{eq:09012026uno}
and
$
\bold{K}^{(2)}(\bold x):
=1_3 1_3^\top+\sum_{j=1}^{\infty}T_j^{(1)}\sigma(\zeta_j^{(1)}(\bold x))\sigma(\zeta_j^{(1)}(\bold x))^\top.
$
Here,  $\{\zeta_j^{(1)}(\bold x)\}_{j\geq 1}$ is a sequence of i.i.d.  r.v.'s with
$\zeta_1^{(1)}(\bold x)\sim\mathcal{N}_3(0,\bold{K}^{(1)}(\bold x))$,  independent of  
$\{T_j^{(1)}\}_{j\geq 1}$,  which are the points of a Poisson process on $(0,\infty)$ with mean measure $\rho$ defined by
\eqref{eq:01072025uno}.  
By Theorem \ref{thm:bayesianlimit} if we denote by $G^{(2)}(\bold x)$ the infinite-width limit of the posterior law of the output,  it holds
$
G^{(2)}(\bold x)\sim\mathcal{GM}_3^{\mathbb S}(\lambda^{(2)}(\bold x,\bold y),\boldsymbol\Lambda^{(2)}(\bold x)).
$
\begin{figure}[!h]
\centering
\includegraphics[width=12.5cm]{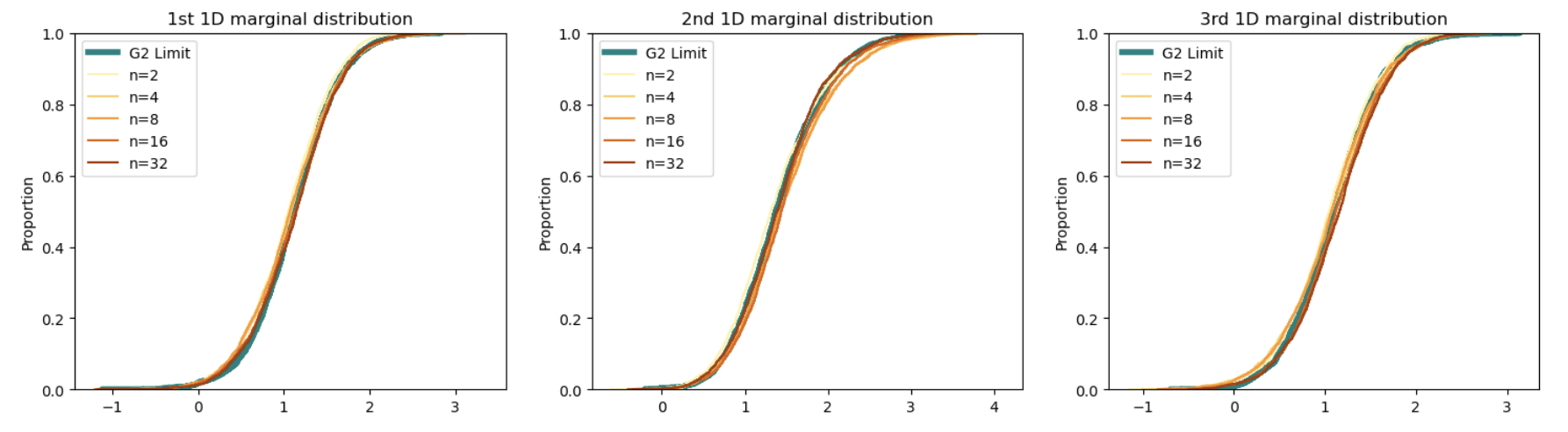}

\caption{{\bf Simulation of Model 2}. The non-green curves are estimates of the one dimensional marginal distribution functions (for different values 
$n_1=n=2,4,8,16, 32$ of the number of nodes in the hidden layer) of the posterior law of the $3$-variate output $Z^{(2)}(\bold x)$.  The green curves are the one dimensional marginal distribution functions of the distribution function of the $3$-variate infinite-width limit $G^{(2)}(\bold x)$.  All
the other parameters of the model are specified in Section~\ref{ss:model2}.}
\label{fig2}
\end{figure}
Therefore,  the probability of the event $\{G^{(2)}(\bold x)\in A\}$,  $A\in\mathcal{B}(\mathbb R^3)$ is equal to
\begin{align}
&\mathbb{E}_{\mathbb S}\left[\int_A\varphi_{(\lambda^{(2)}(\bold x,\bold y),\boldsymbol\Lambda^{(2)}(\bold x))}(\xi)\mathrm{d}\xi\right]
\nonumber\\
&\quad
\propto\mathbb{E}_{{\rm prior}}\left[\frac{
\exp\left(\mathrm{Tr}(\bold y(\bold{Id}_3+(2\bold{K}^{(2)}(\bold x))^{-1})^{-1})\bold y^\top\right)}{(\mathrm{det}(\bold{Id}_3+2\bold{K}^{(2)}(\bold x)))^{n_{2}/2}}
\int_A\varphi_{(\lambda^{(2)}(\bold x,\bold y),\boldsymbol\Lambda^{(2)}(\bold x))}(\xi)\mathrm{d}\xi\right].
\nonumber
\end{align}
So,  we sample from the law of 
$G^{(2)}(\bold x)$
simulating the random matrix $\bold{K}^{(2)}(\bold x)$ under the prior,  and then using a Monte Carlo estimator.  
To sample the random matrix $\bold{K}^{(2)}(\bold x)$ (under the prior),  we note that
letting $\{T_{(j)}^{(1)}\}_{j\geq 1}$,  $T_{(1)}^{(1)}>T_{(2)}^{(1)}>\ldots$,  denote the sequence of points $\{T_j^{(1)}\}_{j\geq 1}$ ordered in decreasing way,  one has
\[
\sum_{j\geq 1}T_j^{(1)}\sigma(\zeta_j^{(1)}(\bold x))\sigma(\zeta_j^{(1)}(\bold x))^\top\stackrel{\mathcal L}{=}\sum_{j\geq 1}T_{(j)}^{(1)}\sigma(\zeta_j^{(1)}(\bold x))\sigma(\zeta_j^{(1)}(\bold x))^\top,
\]
and,  as noticed in Appendix E.3.2 of \cite{LAJLYC},
$
T_{(j)}^{(1)}=\frac{4}{(\sum_{k=1}^{j}E_k)^2},
$
where $\{E_k\}_{k\geq 1}$ is a sequence of i.i.d.  r.v.'s with the exponential law with mean $1$.
If the number of nodes is kept fixed,  we sample from the posterior law of
$Z^{(\ell)}(\bold x)$, $\ell=1,2$,  
simulating these r.v.'s under the prior,  and then using a Monte Carlo estimator as described in Section \ref{ss:model1}.  See Figure~\ref{fig2} for another numerical validation of Theorem \ref{thm:bayesianlimit}. 

\appendix

\section{Proof of Theorem \ref{thm:04122025uno}}\label{AppA}

The proof of Theorem \ref{thm:04122025uno} is based on two lemmas.
The first one,  Lemma \ref{le:28112025},  provides an alternative representation of the neural network; its proof is omitted since it is similar to the proof of Lemma 3 in \cite{MPT}.  The second one,  Lemma \ref{le:30122025due},  is concerned with the weak limit of triangular arrays; its proof can be found in \cite{LAJLYC}, see Corollary 41 and the last lines of the proof of Theorem 16 therein.

Let $\{\widetilde{N}_{jr}^{(\ell)}\}_{j,\ell\geq 1;r\in[d]}$ be a family of i.i.d.  standard normal r.v.'s,  independent of the family $\{V_{n_{\ell-1},j}^{(\ell-1)}\}_{\ell\in[L+1];j\in[n_{\ell-1}]}$.
For $\ell\in[L]$,  we define the symmetric $d\times d$ positive semi-definite random matrices 
\begin{equation}\label{eq:21112025uno}
\bold{H}^{(\ell)}(\bold x)=C_B 1_d 1_d^\top+C_W\sum_{j=1}^{n_\ell}V_{n_{\ell},j}^{(\ell)}\sigma(\gamma_j^{(\ell-1)}(\bold x))\sigma(\gamma_j^{(\ell-1)}(\bold x))^\top
\end{equation}
and
$
\bold{H}^{(0)}(\bold x)=\bold{K}^{(1)}(\bold x):=C_B+\frac{C_W}{n_0}\bold{x}^\top\bold x.
$
Here,  for $\ell\in[L]$ and $j\in[n_\ell]$, 
\[
\gamma_j^{(\ell-1)}(\bold x):=\left(\sum_{t=1}^{d}\widetilde{N}_{jt}^{(\ell)}H_{t1}^{(\ell-1),\sharp}(\bold x),\ldots,\sum_{t=1}^{d}\widetilde{N}_{jt}^{(\ell)}H_{td}^{(\ell-1),\sharp}(\bold x)\right)^\top.
\]
Moreover,  for $\ell=0,\ldots,L$ and $r,s\in[d]$,  we denote by $H_{rs}^{(\ell)}(\bold x)$ the $rs$-entry of
$\bold{H}^{(\ell)}(\bold x)$ and by $H_{rs}^{(\ell),\sharp}(\bold x)$ the $rs$-entry of
 $[\bold{H}^{(\ell)}(\bold x)]^\sharp$ (i.e.,  the square-root of $\bold{H}^{(\ell)}(\bold x)$).  Note that,  given $\bold{H}^{(\ell-1)}(\bold x)$,  $\{\gamma_j^{(\ell-1)}(\bold x)\}_{j\in[n_\ell]}$ are i.i.d.  with law $\mathcal{N}_d(0,\bold{H}^{(\ell-1)}(\bold x))$.
Note also that
\[
H_{rs}^{(\ell)}(\bold x):=C_B+C_W\sum_{j=1}^{n_\ell}V_{n_\ell j}^{(\ell)}\sigma\left(\sum_{t=1}^{d}\widetilde{N}_{jt}^{(\ell)}H_{tr}^{(\ell-1),\sharp}(\bold x)\right)\sigma\left(\sum_{t=1}^{d}\widetilde{N}_{jt}^{(\ell)}H_{ts}^{(\ell-1),\sharp}(\bold x)\right).
\]
If we set $\boldsymbol{\gamma}^{(\ell-1)}(\bold x)=[\gamma_1^{(\ell-1)}(\bold x),\ldots,\gamma_{n_{\ell}}^{(\ell-1)}(\bold x)]\in\mathbb{R}^{d\times n_\ell}$, $\bold{V}^{(\ell)}=\mathrm{diag}\left(V_{n_{\ell},1}^{(\ell)},\ldots,V_{n_{\ell},n_{\ell}}^{(\ell)}\right)\in \mathbb{R}^{n_\ell\times n_\ell}$ and $\widetilde{\bold N}^{(\ell)}=(\widetilde{N}_{ij}^{\ell})\in \mathbb{R}^{n_{\ell}\times d}$, then 
\begin{gather}
\bold{H}^{(\ell)}(\bold x)=C_B 1_d 1_d^\top+C_W\sigma\left(\widetilde{\bold N}^{(\ell)}[\bold{H}^{(\ell-1)}(\bold x)]^\sharp\right)^\top\bold{V}^{(\ell)}\sigma\left(\widetilde{\bold N}^{(\ell)}[\bold{H}^{(\ell-1)}(\bold x)]^\sharp\right)\label{eq:new3}\\
\boldsymbol{\gamma}^{(\ell-1)}(\bold x)=\left(\widetilde{\bold N}^{(\ell)}[\bold{H}^{(\ell-1)}(\bold x)]^\sharp\right)^\top.
\label{eq:new4}
\end{gather}
Recall that the characteristic function of a radom matrix $\bold U$ is denoted by $\phi(\boldsymbol{\theta};\bold U)$. 
\begin{lem}\label{le:28112025}
Under the foregoing assumptions and notation,  it holds
$
\bold{Z}^{(L+1)}(\bold x)\overset{\mathcal L}{=}\widetilde{\bold N}^{(L+1)}[\bold{H}^{(L)}(\bold x)]^\sharp.
$
Hence
\begin{equation}\label{eq:24112025uno}
\phi(\boldsymbol{\theta};\bold{Z}^{(L+1)}(\bold x))=\mathbb E\left[\exp\left(-\frac{1}{2}\mathrm{Tr}\big[\boldsymbol{\theta}^\top\bold{H}^{(L)}(\bold x)\boldsymbol{\theta}\big]\right)\right],
\quad\boldsymbol{\theta}\in\mathbb{R}^{d\times n_{L+1}}.
\end{equation}
\end{lem}

\begin{lem}\label{le:30122025due}
Let $\{V_{n,j}\}_{n\geq 1,j\in[n]}$ be a sequence of non-negative r.v.'s such that
$
\sum_{j=1}^{n}V_{n,j}\overset{\mathcal L}{\to} {\rm ID}(a,\rho),
$
as $n\to\infty$,
where $a\geq 0$ is a non-negative constant and $\rho$ is a L\'evy measure on $(0,\infty)$,  and let $\{\xi_j\}_{j\geq 1}$ be a sequence of i.i.d.  r.v.'s with law $\mathcal{N}_d(0,\bold C)$,  independent of $\{V_{n,j}\}_{n\geq 1,j\in[n]}$.  If $\sigma$ satisfies the assumption $(i)$ of Theorem \ref{thm:starttraining},  then
\[
\sum_{j=1}^{n}V_{n,j}\sigma(\xi_j)\sigma(\xi_j)^\top\overset{\mathcal L}{\to} a\mathbb E[\sigma(\xi_1)\sigma(\xi_1)^\top]+\sum_{j=1}^{N(0,\infty)}T_j\sigma(\xi_j)\sigma(\xi_j)^\top,
\quad\text{as $n\to\infty$}
\]
where $N=\{T_j\}_{j\geq 1}$ is a Poisson process on $(0,\infty)$ with mean measure $\rho$,  independent of $\{\xi_j\}_{j\geq 1}$.
\end{lem}

Before proving Theorem \ref{thm:04122025uno},  we need more details about sequential limits performed in a given order.  For $\ell\in \mathbb{N}^*$,  let $f:\mathbb{N}^{\ell}\rightarrow \mathbb{R}$ be a function,  and,
for $S:=\{i_1,\ldots,i_r\}\subseteq [\ell]$,  $i_1<i_2<\ldots<i_r$, 
let $\beta: S\rightarrow [\ell]$ be an injection.  By $\beta\lim f$ we mean $\lim_{n_{\beta(i_r)}\to\infty}\lim_{n_{\beta(i_{r-1})}\to\infty}\cdots\lim_{n_{\beta(i_1)}\to\infty}f$ and we say that the limit is sequential along $\beta$.  
Let $\pi$ be a permutation of $[\ell]$ and $j\in\{0\}\cup[\ell]$. By $(\pi\preceq j)$ we denote the injection obtained by restricting $\pi$ on $[j]$. Analogously, $(\pi\succeq j)$ denotes the injection obtained by restricting $\pi$ on $\{j,j+1,\ldots,\ell\}$. If $j=0$,  we denote by ($\pi\preceq 0$) the unique injection $\emptyset\to[L]$ and we set $(\pi\preceq 0)\lim f:=f$. Clearly, if $j=\ell$, then $(\pi\preceq j)=\pi$ while $(\pi\succeq j)$ is the unique permutation of $\{\ell\}$. Moreover, if $j\not=\ell$, then  
\begin{equation}\label{eq:seqlim}
\pi\lim f=(\pi\succeq j+1)\lim\left[(\pi\preceq j)\lim f\right].
\end{equation}
The preceding definitions can be extended in a natural way to other notions of limits.  For instance,  if $f$ is random,  we write $f\xrightarrow{\beta\mathcal{L}}\overline{f}$ to mean that $f$ converges in law to $\overline{f}$ sequentially along $\beta$. Analogously, 
we write $f\xrightarrow{\beta\mathbb P}\overline{f}$ to mean that $f$ converges
to $\overline{f}$ $\mathbb P$-a.s.  sequentially along $\beta$.  Hereon,  if $\pi$ is a permutation of $[\ell]$,  and $f\xrightarrow{\pi\mathcal{L}}\overline{f}$,  then we denote by $\overline{f}_{(\pi\preceq j)}$ the weak sequential limit of $f$ along $(\pi\preceq j)$.  

\begin{proof}[Proof of Theorem \ref{thm:04122025uno}]
	 For ease of notation, we omit the explicit reference to the input matrix $\bold x$ in every symbol it occurs. Since $\textrm{vec}$ and $\top$ are isometries of Euclidean spaces, we equivalently prove the theorem for the random matrices $\bold{Z}^{(L+1)}$ and $\bold{G}^{(L+1)}$ rather that for the r.v.'s $Z^{(L+1)}$ ad $G^{(L+1)}$.  Let $\mathcal{S}_+^d$ be the space of the order $d$ symmetric and positive semi-definite real matrices endowed with the Frobenius norm,  and 
note that,  for any fixed $\boldsymbol{\theta}\in\mathbb{R}^{d\times d}$,  the mapping $
\mathcal{S}_+^d\ni\bold A\mapsto \exp\left(-\frac{1}{2}\mathrm{Tr}\big[\boldsymbol{\theta}^\top\bold{A}\boldsymbol{\theta}\big]\right)
$ is real-valued,  continuous and bounded.  Hence,  by \eqref{eq:24112025uno} and the definition of weak convergence it holds 
\[
\bold{H}^{(L)}\xrightarrow{\pi\mathcal{L}}\bold{K}^{(L+1)}\Longrightarrow 
\pi\lim\phi(\boldsymbol{\theta};\bold{Z}^{(L+1)})=\mathbb E\left[\exp\left(-\frac{1}{2}\mathrm{Tr}\big[\boldsymbol{\theta}^\top\bold{K}^{(L+1)}\boldsymbol{\theta}\big]\right)\right]=\phi(\boldsymbol{\theta};\bold{G}^{(L+1)}).
\]
Therefore to prove the theorem it suffices to prove that $\bold{H}^{(L)}\xrightarrow{\pi\mathcal{L}}\bold{K}^{(L+1)}$,  for every permutation $\pi$ of $[L]$.
We proceed by induction on the number of layers $L$.  The basis of the induction: $\bold{H}^{(1)}\xrightarrow{\pi\mathcal{L}}\bold{K}^{(2)}$,  reduces to
\begin{equation}\label{eq:28112025basis}
\bold{H}^{(1)}{\xrightarrow[n_1\to\infty]{\mathcal{L}}}\bold{K}^{(2)}
\end{equation}
since,  for $L=1$,  the unique permutation of $[L]$ is the identity.  We defer the proof of \eqref{eq:28112025basis} after the proof of the inductive step.  So,  let $L\geq 3$ and assume
\begin{equation}\label{eq:26112025tre}
\bold{H}^{(L-1)}\xrightarrow{\tau\mathcal{L}}\bold{K}^{(L)}\quad\text{for every permutation $\tau$ of $[L-1]$}.
\end{equation}
Clearly,  by the continuity theorem the claim follows if we prove
\begin{equation}\label{eq:new6}
	\pi\lim\phi(\boldsymbol{\theta};\bold{H}^{(L)})=\phi(\boldsymbol{\theta};\bold{K}^{(L+1)})\quad\text{$\forall \boldsymbol{\theta}\in\mathbb{R}^{d\times d}$ and every permutation $\pi$ of $[L]$},
\end{equation}
assuming \eqref{eq:28112025basis} (to be proved) and \eqref{eq:26112025tre}.
Let $\pi$ be any permutation of $[L]$ and let $j^*=\pi^{-1}(L)$. Hence $(\pi\preceq j^*-1)=(\tau\preceq j^*-1)$ for some permutation $\tau$ of $[L-1]$. Since the mapping $\bold A\to \bold A^\sharp$ is continuous on $\mathcal{S}_+^d$,  the inductive step \eqref{eq:26112025tre} yields $
[\bold{H}^{(L-1)}]^\sharp\xrightarrow{(\pi\preceq j^*-1)\mathcal{L}}\bold K^{(L)}_{(\pi\preceq j^*-1)}$. Since, by construction, $\bold{H}^{(L-1)}$ is independent of $\bold{V}^{(L)}$ and $\widetilde{\bold N}^{(L)}$ and $\sigma$ is continuous, by \eqref{eq:new3} it follows that
\[
\sigma\left(\widetilde{\bold N}^{(L)}[\bold{H}^{(L-1)}]^\sharp\right)^\top\!\bold{V}^{(L)}\sigma\left(\widetilde{\bold N}^{(L)}[\bold{H}^{(\ell-1)}]^\sharp\right)
\]
converges to
\[
\sigma\left(\widetilde{\bold N}^{(L)}[\bold{K}^{(L)}_{(\pi\preceq j^*\!-\!1)}]^\sharp\right)^\top\bold{V}^{(L)}\sigma\left(\widetilde{\bold N}^{(L)}[\bold{K}^{(L)}_{(\pi\preceq j^*\!-\!1)}]^\sharp\right)
\]
in law along ${(\pi\preceq j^*\!-\!1)}$.
Note that also $[\bold{K}^{(L)}_{(\pi\preceq j^*-1)}]^\sharp$
is independent of $\bold{V}^{(L)}$ and $\widetilde{\bold N}^{(L)}$. Therefore, if $\bold{\Xi}:=[\xi_1,\ldots,\xi_{n_L}]\in \mathbb{R}^{d\times n}$ is the random matrix defined by $\bold{\Xi}=\widetilde{\bold N}^{(L)}[\bold{K}^{(L)}_{(\pi\preceq j^*\!-\!1)}]^\sharp$, then $\bold{\Xi}$ is independent of $\bold{V}^{(L)}$ and,  given $[\bold{K}^{(L)}_{(\pi\preceq j^*\!-\!1)}]^\sharp$,
$\xi_1,\ldots,\xi_{n_L}$ are i.i.d.  $d$-dimensional r.v.'s
with law $\mathcal{N}_d(0,\bold{K}^{(L)}_{(\pi\preceq j^*\!-\!1)})$.  Consequently, in view of \eqref{eq:21112025uno}, one has
\begin{align*}
(\pi\preceq\!j^*\!-\!1)\lim\phi(\boldsymbol{\theta};\bold{H}^{(L)})=\mathbb E\left[\exp\left(\bold{i}\langle\boldsymbol\theta,C_B\bold{1}_d\bold{1}_d^\top+C_W\sum_{j=1}^{n_L}V_{n_L j}^{(L)}\sigma(\xi_j)\sigma(\xi_j)^\top\rangle_F\right)\right].\nonumber
\end{align*}
On the other hand,  with $n_{\pi(j^*)}=n_L$, by Lemma \ref{le:30122025due}
we have
\[
C_B\bold{1}_d\bold{1}_d^\top+C_W\sum_{j=1}^{n_L}V_{n_L j}^{(L)}\sigma(\xi_j)\sigma(\xi_j)^\top{\xrightarrow[n_L\to\infty]{\mathcal{L}}} \bold{K}^{(L+1)}_{(\pi\preceq j^*)}
\]
where, referring to \eqref{eq:14052025due}, $\bold{K}^{(L+1)}_{(\pi\preceq j^*)}$ is defined similarly to $\bold{K}^{(L+1)}$ but with the $\xi_j$'s in place of the $\zeta_j^{(L)}$'s and $\bold{K}^{(L)}_{(\pi\preceq j^*\!-\!1)}$ in place of $\bold{K}^{(L)}$,  i.e.,
\begin{align}
\bold{K}^{(L+1)}_{(\pi\preceq j^*)}:=C_B 1_d 1_d^\top+C_W\Biggl(a^{(L)}\mathbb{E}[\sigma(\xi_1)\sigma(\xi_1)^\top\,|\,\bold{K}^{(L)}_{(\pi\preceq j^*\!-\!1)}]+\sum_{j=1}^{N_{L}((0,\infty))}T_j^{(L)}\sigma(\xi_j)\sigma(\xi_j)^\top\Biggr).\label{eq:27112025quattro}
\end{align}
%
Since $L=\pi(j^*)$ we have proved
\begin{equation}\label{eq:new5}
	(\pi\preceq j^*)\lim\phi(\boldsymbol{\theta};\bold H^{(L)})=\lim_{n_L\to\infty}\left[(\pi\preceq\!j^*\!-\!1)\lim\phi(\boldsymbol{\theta};\bold{H}^{(L)})\right]=
	\mathbb E\left[\exp\left(\bold{i}\mathrm{Tr}\left[\boldsymbol{\theta}^\top
	\bold{K}^{(L+1)}_{(\pi\preceq j^*)}
\right]
	\right)\right].
\end{equation}
If $j^*=L$, then we are done. Indeed, since $(\pi\preceq j^*)=\pi$ and $(\pi\preceq j^*\!-\!1)=\tau$ for some permutation $\tau$ of $[L-1]$, it follows that $\bold{K}^{(L)}_{(\pi\preceq j^*\!-\!1)}=\bold{K}^{(L)}$ by \eqref{eq:26112025tre}. Hence $\bold{K}^{(L+1)}_{(\pi\preceq j^*)}=\bold{K}^{(L+1)}$ by \eqref{eq:27112025quattro}, after recalling \eqref{eq:14052025due}. We thus assume $j^*\not=L$ and prove that $\bold{K}^{(L+1)}_{(\pi\preceq j^*)}\xrightarrow{(\pi\succeq j^*+1)\mathcal{L}}\bold{K}^{(L+1)}$. Since $(\pi\succeq j^*\!+\!1)=(\tau\succeq j^*\!+\!1)$ for some permutation $\tau$ of $[L]$, it follows that $\bold{K}^{(L)}_{(\pi\preceq j^*)}\xrightarrow{(\pi\succeq j^*+1)\mathcal{L}}\bold{K}^{(L)}$ by \eqref{eq:26112025tre} . A simple computation with characteristic functions shows
\begin{equation}\label{eq:27112025uno}
(\xi_1,\ldots,\xi_{n_L})\xrightarrow{(\pi\succeq j^*+1)\mathcal{L}}
(\zeta_{1}^{(L)},\ldots,\zeta_{n_L}^{(L)}).
\end{equation}
By \eqref{eq:27112025uno} and Skorohod's representation theorem,  there exists a probability space,  say $(\tilde{\Omega},\tilde{\mathcal{F}},\tilde{\mathbb P})$,  
and copies of the r.v.'s $\xi$ and $\zeta$ (still denoted, by a little abuse of notation,  $\xi$ and $\zeta$,  respectively),  where the convergence is $\tilde{\mathbb P}$-a.s.  Recall that if $\{X_n\}_{n\geq 1}$ is a sequence of $d$-dimensional Gaussian r.v.'s and $X$ is a $d$-dimensional Gaussian r.v. ,  one has that if $X_n\xrightarrow{\mathcal{L}}X$,  then,  for any $p_1,\ldots,p_k\geq 0$,
$0\leq k\leq d$,  $\{i_1,\ldots,i_k\}\subseteq [d]$ it holds
$
\mathbb E|X_n(i_1)|^{p_1}\ldots|X_n(i_k)|^{p_k}\to\mathbb E|X(i_1)|^{p_1}\ldots|X(i_k)|^{p_k}.
$ 
Since $\sigma$ is continuous and is either bounded or with subpolynomial growth,  using the dominated convergence theorem in the bounded case and the generalized dominated convergence theorem in the case of a subpolynomial growth,  one has
\[
\tilde{\mathbb{E}}\left[\sigma(\xi_1)\sigma(\xi_1)^\top\,|\,\bold{K}^{(L)}_{(\pi\preceq j^*)}\right]\xrightarrow{(\pi\succeq j^*+1)\tilde{\mathbb P}}\tilde{\mathbb{E}}\left[\sigma(\zeta_{1}^{(L)}\sigma(\zeta_{1}^{(L)})^\top\,|\,\bold{K}^{(L)}\right].
\]
Using the continuity of $\sigma$,  one also has
\[
\sum_{j=1}^{N_{L}((0,\infty))}T_j^{(L)}\sigma(\xi_j)\sigma(\xi_j)^\top\xrightarrow{(\pi\succeq j^*+1)\tilde{\mathbb P}}\sum_{j=1}^{N_{L}((0,\infty))}T_j^{(L)}\sigma(\zeta_{j}^{(L)})\sigma(\zeta_{j}^{(L)})^\top
\]
where,  with a little abuse of notation,  we still denote by $N_L=\{T_j^{(L)}\}$ a copy of $N_L=\{T_j^{(L)}\}$ on $\tilde\Omega$.  Considering \eqref{eq:05122025due},
this convergence is clear if $\rho^{(L)}((0,\infty))<\infty$.  Otherwise,  assumption \eqref{eq:05122025uno} holds and the convergence is a consequence of the dominated convergence theorem.  Indeed,  being $\rho^{(L)}$ a L\'evy measure,  one has that the versions on $\tilde\Omega$ of 
the r.v.'s
$\sum_{j=1}^{N_{L}((0,1))}T_j^{(L)}$ and $\sum_{j:\,T_j^{(L)}\in[1,\infty)}T_j^{(L)}$
are finite $\tilde{\mathbb P}$-a.s.  (the first sum has a finite mean and the second sum has a finite number of addends).  So,  by \eqref{eq:27112025quattro},  the version of $\bold{K}^{(L+1)}_{(\pi\preceq j^*)}$ on $\tilde{\Omega}$ converges
to the version of $\bold{K}^{(L+1)}$ on $\tilde{\Omega}$,  $\tilde{\mathbb P}$-a.s.. Therefore,  by the dominated convergence theorem, in view of \eqref{eq:seqlim} and \eqref{eq:new5}, one has
\[
\begin{split}
\pi\lim\phi(\boldsymbol{\theta};\bold{H}^{(L)})&=(\pi\succeq j^*+1)\lim\left[(\pi\preceq j^*)\lim\phi(\boldsymbol{\theta};\bold H^{L})\right]\\
&=(\pi\succeq j^*+1)\mathbb E\left[\exp\left(\bold{i}\mathrm{Tr}\left[\boldsymbol{\theta}^\top
\bold{K}^{(L+1)}_{(\pi\preceq j^*)}\right]
\right)\right]
=\mathbb E\left[\exp\left(\bold{i}\mathrm{Tr}\left[\boldsymbol{\theta}^\top
\bold{K}^{(L+1)}\right]\right)\right]
\end{split}
\]
and \eqref{eq:new6} is proved. It remains to prove \eqref{eq:28112025basis}. By construction $\bold{H}^{(0)}=\bold{K}^{(1)}$. Hence, with $\ell=L=1$, by \eqref{eq:new4}, for $j\in[n_1]$,  it holds that $\gamma_j^{(0)}\overset{\mathcal L}{=}\zeta_{j}^{(1)}$, where $\zeta_{j}^{(1)}\sim\mathcal{N}_d(0,\bold{K}^{(1)})$. Consequently, by \eqref{eq:21112025uno}, one has
\[
\bold{H}^{(1)}=C_B 1_d 1_d^\top+C_W\sum_{j=1}^{n_1}V_{n_1 j}^{(1)}\sigma(\zeta_{j}^{(1)})\sigma(\zeta_{j}^{(1)})^\top.
\]  
Hence, after recalling \eqref{eq:14052025due}, by Lemma \ref{le:30122025due}, $\bold{K}^{(2)}$ is the limit in law of $\bold{H}^{(1)}$ as $n_1\to\infty$. The proof is thus completed.  
\end{proof}








\end{document}